\documentclass[10pt]{article}

\usepackage{lineno,hyperref}
\modulolinenumbers[5]

\usepackage{fancybox}
\usepackage[tight,normalsize,sf,SF]{subfigure}
\usepackage{multirow}
\usepackage{amsmath}
\usepackage{fixltx2e}
\usepackage{array}
\usepackage{booktabs}
\usepackage{xcolor}
\usepackage{rotating}
\usepackage[british]{babel}
\usepackage{url}
\usepackage{color,soul}
\usepackage{csvsimple}
\usepackage{blindtext}
\usepackage[colorinlistoftodos]{todonotes}
\usepackage{graphicx}

\graphicspath{{./images/}}

\begin{document}
\newcommand\NCD{\mathit{NCD}}
\newcommand\DSC{\mathit{DSC}}

\DeclareRobustCommand{\hlmag}[1]{{\sethlcolor{magenta}\hl{#1}}}
\title{Algorithmic Clustering based on String Compression to Extract P300 Structure in EEG Signals\thanks{This is the postprint version of an article published in \emph{Computer Methods and Programs in Biomedicine} (Elsevier). The final authenticated version is available online at \href{https://doi.org/10.1016/j.cmpb.2019.03.009}{https://doi.org/10.1016/j.cmpb.2019.03.009}.}}

\date{}

\author{Guillermo Sarasa, Ana Granados, Francisco B Rodr\'iguez}

\maketitle 

% \author{Guillermo Sarasa, Ana Granados   and Francisco de Borja Rodr\'iguez% <-this % stops a space
% \IEEEcompsocitemizethanks{\IEEEcompsocthanksitem A. Granados is with the CES Felipe II, Universidad Complutense de Madrid, Aranjuez, Spain. \protect\\
% E-mail: ana.granados@ajz.ucm.es
% \IEEEcompsocthanksitem G. Sarasa and F. Rodr\'iguez are with the Escuela Polit\'ecnica Superior, Universidad Aut\'onoma de Madrid, Madrid, Spain. \protect\\
% E-mail: guillermo.sarasa@predoc.uam.es, f.rodriguez@uam.es}% <-this % stops a space
% \thanks{Manuscript received xxxx; revised xxxx.}}

\begin{abstract}

\textit{Background and objectives:} P300 is an Event Related Potential control signal
widely used in Brain Computer Interfaces. Using the oddball paradigm, a P300 speller allows a
human to spell letters through P300 events produced
by  his/her brain. One of the most common issues in the detection of this event is that its
structure may differ between different subjects and over time for a specific
subject. The main purpose of this work is to deal with this inherent variability
and identify the main structure of P300 
using algorithmic clustering based on string compression.

\textit{Methods}: In this work, we make use of the Normalized Compression Distance (NCD)
to extract the main 
structure of the signal regardless of its inherent variability. In order 
to apply compression distances,
we carry out a novel signal-to-ASCII process that transforms and merges different events into
suitable objects to be used by a compression algorithm.
Once the ASCII objects are created, we use NCD-driven clustering as a tool to
analyze if our object creation method suitably represents the information
contained in the signals and to explore if compression distances are a valid
tool for identifying P300 structure. With the purpose of increasing the level
of 
generalization of our study, we apply two different clustering methods: a
hierarchical clustering algorithm based on the minimum quartet tree method and
a multidimensional projection method.

\textit{Results:}
Our experimental results show reasonable clustering performance over different
experiments, showing the structure-extraction capabilities of our procedure. Two
datasets with recordings in different scenarios were used to analyze the problem
and validate our results, respectively. It has to be pointed out that when the
clustering performance over individual electrodes is analyzed, higher P300
activity is found in similar regions to other articles using the same data. This
suggests that our approach might be used as an electrode-selection criteria.

\textit{Conclusions}: The proposed NCD-driven clustering methodology can be used to
discover the structural characteristics of EEG and thereby, it is suitable as a
complementary methodology for the P300 analysis. 

\end{abstract}


\section{Introduction}\label{Introduction}

\begin{figure}
  \centering
  \includegraphics[width=0.75\textwidth]{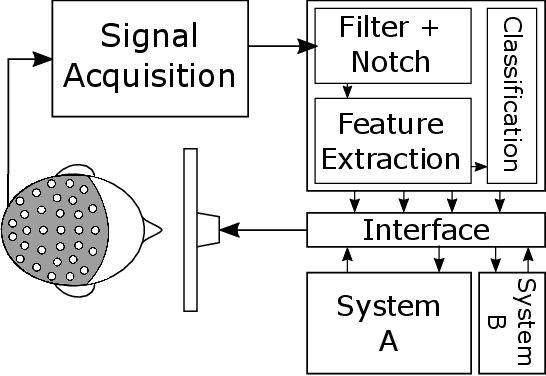}
  \caption{Scheme of a BCI stimuli-based-on system. The signal is recorded   by
    means of certain acquisition technique, for instance the EEG recording. Then
    the digital signal obtained is processed (filtered 
    + feature extraction + classification) in order to be parsed into commands.
    Finally, this extracted information (the commands) will be send through an
    interface with the different systems or devices to do the desired actions
    (system A and/or system B in figure). These systems can be for instance a
    commanded wheelchair, a BCI speller to send messages, a commanded robotic
    arm, etc.} 
  \label{fig:systemb}
\end{figure}

Brain Computer Interfaces (BCIs) offer a human-computer interaction, translating
patterns of brain activity into commands to perform different tasks
\cite{rao_2013,amiri_review_2013}. BCIs have been
used, mostly, by people with motor disabilities, as an external control system,
for example, to control a wheelchair or to speak through an artificial voice.
The general architecture of a BCI can be seen in Figure \ref{fig:systemb}. First,
brain activity is recorded by a (previously defined) signal
acquisition technique  which contains a pattern as a control signal for the BCI.
Next, the signal is filtered and follows several feature extraction methods
that aim to improve the BCI precision. Finally, a classifier is trained to
discriminate between patterns and communicate, through an interface, with the
different systems. Although this architecture can differ between BCI
systems, they are usually defined by the acquisition technique and the control
signal used.

One of the most commonly used acquisition techniques in the development of BCIs
is Electroencephalogram \cite{arroyo_functional_1993} (EEG). This signal is
composed of measurements of brain activity which may contain useful information
about the brain state but their non-linear and non-stationary nature makes them
very difficult to analyze
\cite{krusienski_critical_2011,nicolas-alonso_brain_2012,MCFARLAND2017194}. 
Despite this, EEG signals are commonly used in the context of BCI problems
due to their time resolution, their low-cost and the fact that they can be
easily and non-invasively acquired through electrodes placed on the scalp.

The possibilities and interaction of EEGs are defined by the control
signal. Several control signals have been used in the development of BCIs,
being the most commonly used in EEG, the so-called Event Related Potentials
(ERPs). These 
are defined as a brain electrophysiological response (very small positive or
negative deflections of voltages generated in the brain structures) as a
consequence of an external stimulus. Examples of such control signals are the
Steady State Visual Evoked Potential \cite{morgan_selective_1996} (frequency
response to visual stimulation), the Sensorimotor Rhythms
\cite{arroyo_functional_1993, yuan_brain-computer_2014} (subject's will to
imagine movements), or the P300 \cite{picton_p300_1992,gao_visual_2014}
(response to visual, auditory or tactile stimulation). The first two control
signals require previous user training while the latter does not. This, in
addition to the variety of applications of P300
\cite{kaufmann_toward_2014,halder_training_2016,fukami_robust_2016,chen_enhancing_2017},
makes P300s very appealing to be used to develop BCIs. This is the reason why,
in this work, we have focused on the P300 control signal.

Typically, a P300 is
defined as a positive amplitude increment in the EEG signal due to an infrequent
stimulus \cite{picton_p300_1992}. Particularly, it is defined as a positive peak in the
recorded signal 300 ms after the apparition of a relevant stimulus. P300 is a very
complex wave defined by several structural components, some of which have been
identified in the literature
\cite{blankertz_single-trial_2011,chen_enhancing_2017}. Some examples 
of these structural components are the previous positive deflection potentials, P100
and P200, and the previous negative deflection potentials, N100 and N200
\cite{light_validation_2015,ouyang_exploiting_2017}.
The latency and amplitude of these components, and many others, may differ
making the P300-ERP difficult to identify. Some works have studied this
variability \cite{fira_comparison_2016} and how these
factors can vary between subjects (inter-subject variability)
\cite{van_dinteren_p300_2014} and over time for a specific subject
(intra-subject variability)
\cite{lu_effects_2013,ouyang_exploiting_2017}.

\begin{figure}
  \centering
  \includegraphics[width=0.85\textwidth,trim={1cm 0.cm 1.5cm 1.75cm},clip]{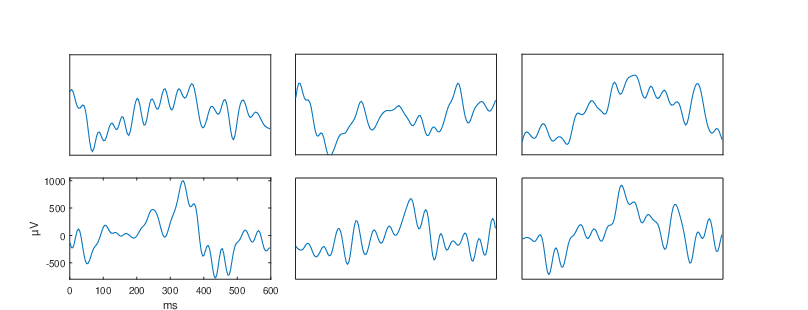}
  \caption{Scheme of six P300-ERPs taken from the II BCI Competition problem 2b (speller matrix) dataset. Each interval, of 600ms, was taken after a stimulus was shown to the subject and the intensified row or column belong to the target character. The P300-ERP should manifest an increment of amplitude around 300ms after the stimulus.}
  \label{fig:6p300}
\end{figure}

In order to give an insight into how complex and
variable the structure of P300 is, six P300 patterns obtained from the same
subject are depicted in Figure \ref{fig:6p300}. One can observe that although
the six samples correspond to P300 generated by the same person, each of them
has particular structures that makes them very different.

On the other hand,
there are other issues that may appear in BCI systems: (i) variety between
signal sources, (ii) recording of undesired processes or (iii) distortion from
the initial acquisition. These reasons, among others, have made that P300s'
complex structure had been widely defined and studied.

In this work, we deal with the extraction of P300-ERPs in EEG signals by
applying algorithmic clustering based on string compression
\cite{cilibrasi_clustering_2005}. In particular, we focus on one of the most successfully
applied compression 
distances: the Normalized Compression Distance (NCD)\cite{li_similarity_2004}.
This measure has been 
successfully and widely used due to its parameter-free nature, wide
applicability and leading efficacy in numerous domains. Compression distances
are powerful tools that allow identification and capture of complex-data
structure thanks to the ability that compression algorithms have to analyze data
structure in a parameter-free manner. Therefore, we think that compression-based
distances could be a 
useful tool that can help identify P300-ERPs despite their inherent variability.
However, in our opinion, the way in which the signal is coded will be crucial
for the NCD ability to identify relevant P300 structures. Therefore, we propose
a novel signal-to-ASCII process that transforms P300 signals into ASCII objects
to help compression algorithms capture P300 structure more accurately.
Then, we use clustering as a tool to analyze if our object creation method
suitably represents the information contained in the signals and to explore if
compression distances are a valid tool for identifying P300. Summarizing, in
this work we show how to use clustering based on string compression, in the
context of BCI, as a P300 identification methodology.

The rest of the paper is structured as follows. Section
\ref{subsec:compressors} presents compression distances and their common
application research areas. Section \ref{sec:methods} describes the data sets
together with the procedures followed to analyze and transform the signal into data
objects and the clustering methods performed in our experiments. Section \ref{sec:results}
shows our results and some examples of the 
experiments carried out over the proposed object format and two different datasets, for
analysis and validation, respectively. Section \ref{sec:discussion} presents a
briefly review and discussion of our work and the results obtained from the
experiments. Finally, Section \ref{sec:conclusions} summarizes the main
conclusions of our paper.

\vspace{1cm}

\section{Compression distances}
\label{subsec:compressors}

In this work, we propose to use data compression as a tool to
distinguish P300 patterns from non-P300 
patterns in EEG signals. This can be done thanks to the existence of compression
distances, which measure the 
distance or dissimilarity of two objects using compression algorithms, as described below. 

A natural measure of similarity assumes that two objects are similar if the
basic blocks of one are part of 
the other and vice versa. If this happens we can
describe one of the objects by
making reference to the 
blocks belonging to the other one. Thus, the description of an object will be
very simple using the 
description of the other one \cite{cilibrasi_clustering_2005}. Compression algorithms can be used to measure the
dissimilarity of two objects by applying this idea. 

A lossless compressor is a particular case of coding theory in which a
mapping is normally defined through a 
prefix code \cite{Salomon10}. This resultant code has a smaller size than the original code. In this way, the compression
algorithm works reducing the redundancy in a file by searching for information shared in the whole file.
Therefore, a compression algorithm could be used to figure out how redundant two files are if we concatenate
them and make the compressor compress the concatenated file. That is, given two files $x$ and $y$, and their
concatenated file $xy$, if a compression algorithm tried to compress $xy$, it would search for
information shared by both files ($x$ and $y$) in order to reduce the redundancy of $xy$. If the
result were small compared with the compression of each file individually, it would mean that the information
contained in $x$ could be used to code $y$.

This was studied by \cite{cilibrasi_clustering_2005,Li04}, giving rise to the concept of \emph{Normalized Compression Distance}
(NCD), whose mathematical formulation is as follows:

% \%%vspace{-0.2cm}
\begin{eqnarray}
\NCD(x,y)=\frac{\max\{C(xy)-C(x),C(yx)-C(y)\}}{\max\{C(x),C(y)\}}~,\nonumber
\end{eqnarray}
% \%%vspace{-0.2cm}

\noindent where $C$ is a compression algorithm, $C(x)$ is the size of the $C$-compressed version of $x$, $C(xy)$ is the
compressed size of the concatenation of $x$ and $y$, and so on. In practice, the NCD is a non-negative number $0 \leq r
\leq 1 + \varepsilon$ representing how different two objects are. Smaller numbers represent more similar objects. The
$\varepsilon$ in the upper bound is due to imperfections in compression techniques, but for most standard compression
algorithms, one is unlikely to see an $\varepsilon$ above 0.1 \cite{cilibrasi_clustering_2005}.

The NCD has been applied to numerous research
areas because of its parameter-free
nature, wide applicability and leading efficacy. For example, among others, it
has been applied to data mining \cite{cilibrasi07}, profiling diabetes behaviors 
\cite{contreras_profiling_2016},  music clustering 
\cite{GonzalezPardo10,meredith_compression-based_2014},
document clustering \cite{granados_discovering_2015,granados_reducing_2011},
image analysis \cite{Cohen09}, bird song species
classification \cite{sarasa_approach_2017,sarasa_automatic_2018} or video
activity recognition 
\cite{sarasa_compression-based_2018}.
Also, this measure has
  proved to have a high noise resistance \cite{Cebrian07} which makes
  it very appealing for our problem.

To the best of our
knowledge, clustering based on string compression has not been applied to the
extraction of P300 in EEGs yet. The closest works based on compression
algorithms are \cite{berek_classification_2014,fira_size_2017}.
In the former, the authors make use of a quantization technique (Vector Quantization
\cite{noauthor_vector_nodate} which can be seen as a loss compression mapping),
to map the EEG signal into a compressible format. Then, they measure the NCD
between the patterns to classify finger movements. The latter makes use of
Compressed Sensing, to cast the EEG signals into dictionaries to classify
P300-ERPs. There are other works
\cite{alagoz_spiral_2018,kannathal_entropies_2005,bonmati_brain_2017,baravalle_discriminating_2018,gibson_entropy_2018}
that suggest the use of information-based methodologies to extract information
from the signal. In contrast, we reduce the information parsing the signal into ASCII
objects and using the NCD, as dissimilarity measure, to identify clusters of
similar information among the P300-ERPs.

\section{Materials and methods}
\label{sec:methods}

In this section we show the data and methods that have been used to perform the
different experiments presented in
Section \ref{sec:results}. First, we describe the datasets used in our work. Second, we explain how the recorded signals have
been split to create two kinds of data objects: objects that contain P300-ERPs
and objects without P300-ERPs. Third, we describe the clustering algorithms that
have been used in our work. Finally, we explain 
how the clustering quality has been evaluated.

\subsection{Data sets}
\label{subsec:dataset_met}

\begin{figure}
  \centering
  \includegraphics[width=0.75\textwidth]{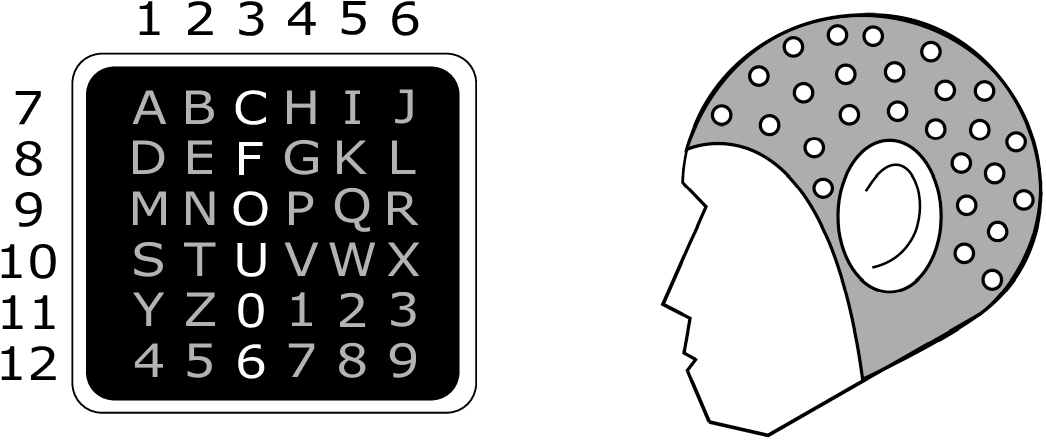}
  \caption{Scheme of a P300 BCI system.  The subject pays attention to a speller matrix program in a screen, in which the different columns and rows intensifies randomly.  At the same time, the brain activity is recorded through EEG. As a result of row or column intensification of the focused character (marked as * in figure \ref{fig:data-structure}) the signal should increment its amplitude. This increment appears after 300ms and does not last longer that 600ms.}
  \label{fig:system}
\end{figure}

We have used the data sets of the second BCI Competition \cite{bci_ii} (Dataset IIb
\cite{noauthor_2nd_nodate}) and third BCI Competition \cite{bci_iii}
(Data set II). We have selected these data sets to analyze and
study the data through compression algorithms, and to validate our results,
respectively. The objective of these 
competitions is to identify the characters that a subject spells with a P300
speller \cite{nam_brain-computer_2018}. As it is shown in 
figure \ref{fig:system}, a $6 \times 6$ matrix of characters is presented to a
subject. The user is asked to focus attention on a sorted set of characters, one
at a time, while different rows and columns are
intensified in a random order. For each character to spell, all the rows and
columns are intensified, but 
only 2 of the 12 ($6$ rows and $6$ columns) intensifications should evoke a P300
response in the subject (those that match to the row and the column where the
desired character to spell is located). From now on, we will refer to these 12
intensifications as \textit{trial}. 
This process, or \textit{trial}, is repeated 15 
times for each character to spell. That is, for a single character, there would
be $12 \times 15$ 
intensifications, where 12 corresponds to the number of intensifications
required to intensify the 6 rows and 
the 6 columns of the matrix, and 15 corresponds to the number of repetitions.

\begin{figure}
  \centering
  \includegraphics[width=0.95\textwidth,trim={0.5cm 1cm 0.8cm 0.55cm},clip]{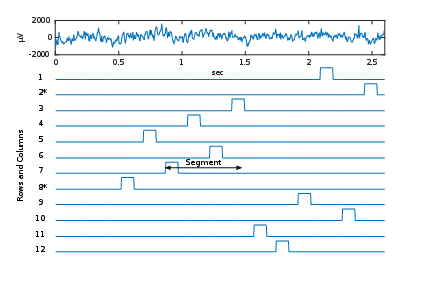}
  \caption{Structure used in the recording session showed in figure
    \ref{fig:system}. This figure is a reduced but more detailed version of
    figure \ref{fig:6p300}. Each numbered source correspond to each stimulus
    (rows 1-6, and columns 7-12) of the speller matrix, where each pulse
    corresponds to an intensification of that row or column. Once the infrequent
    stimulus appears, an amplitude increment should appear in the recorded
    signal 300 ms. The change generated by the ERP should not continue 600ms
    after the stimulus. After the marked row and column should manifest a P300-ERP
    in the figure. In this case, the subject was paying attention to the
    character ``E'' according to figure \ref{fig:system}.} 
  \label{fig:data-structure}
\end{figure}

In figure \ref{fig:data-structure}, we show a sample of the dataset of 2.5
seconds long to show how complex the P300-ERP structure is. In this example, the 
P300 should manifest as a positive increment of amplitude approximately between
0.5 and 1.0 seconds (following the standard definition of the P300-ERP
\cite{picton_p300_1992}). Looking closely to this 
interval, it is possible to glimpse this 
increment of amplitude. Keep in 
mind, however, that in some cases the increment of amplitude is not visible at
first sight. This is caused 
by the signal inherit noise and variability. 

The dataset of the second BCI Competition (IIb problem) comprises three
sessions, but only two of them are 
labeled. In our work, we only use the labeled ones (called session 10 and
session 11 in the dataset). Together both sessions include 42
different characters. As mentioned before, for each character to spell, there
will be 180 ($12 \times 
15$) intensifications. For each intensification, an
interval of 600 ms (144 samples 
at 240 Hz) is recorded in each channel for a total of 64 electrodes. Hereafter,
each of these intervals of 600 
ms will be referred to as \emph{segment} (see figure \ref{fig:data-structure}).
Each of these signal segments is used to create two kinds of 
objects: those that contain P300 events and those that do not contain this type
of event. It is important to notice that,
at each session, the characteristics of the recorded data can
change for several reasons, being the most important ones: (i) 
relocation of the electrodes, (ii) differences in the mental state of the
subject, and (iii) adaptation of the subject as
new experiments are performed. For these reasons, we 
performed different experiments for each session, both separately and jointly.

Besides, in order to increase the level of generalization of our results, we have used another dataset,
particularly the third BCI Competition, Dataset II. The objective of this
competition is also to identify characters spelled by a subject with a P300
speller. However, in this case, two subjects were asked to spell a total of 85
characters instead of 42. A very interesting point is that the quality of the recorded EEG signal is
fairly worse than the quality of the signal from the second BCI Competition (IIb
problem). This fact is given
by the score obtained by the winners of both competitions. While in the
second competition there is a tie (5 people) in the first position with a 1.0
\cite{blankertz_bci_2004} 
classification rate, in the third competition the maximum score is below 0.96
\cite{rakotomamonjy_bci_2008}, 
followed by a 0.9 \cite{blankertz_bci_2006}. Using the third BCI competition
dataset allows us to validate the capabilities of our method with a signal
of worse quality. Thus, we expect to find a decrease in the clustering
quality in contrast with the BCI Competition II. 

\subsection{Signal processing: Object definition}
\label{subsec:signalprocess}

Our experience working with the NCD in different research areas
\cite{sarasa_compression-based_2018,sarasa_automatic_2018,granados_reducing_2011,gonzalez-pardo_influence_2010}
tells us that
the way in which the signal is coded is crucial for the NCD ability to
identify relevant P300 structures
\cite{granados_discovering_2015,granados_is_2012}. Therefore, we process the EEG
signal parsing 
it into objects through the following process: First, the signal is band passed
between 0.5 and 10 Hz (following \cite{rakotomamonjy_bci_2008}, through a
Butterworth filter) and standardized to minimize the undesired components. Then,
the segments from filtered and standardized data are extracted and stored in
vectors to be later labeled as P300 or non-P300, depending on whether the
stimulus corresponds to the intensification of the target row/column or not. As
mentioned above, for every stimulus showed to the subject, a 600ms interval is
stored (see figure \ref{fig:data-structure}). Please note that since there are a
total of 6 rows and 6 columns, there will be five times more objects in the
non-P300 class than in the P300 class.

\begin{figure}[h!]
  \centering
  \includegraphics[width=0.55\textwidth]{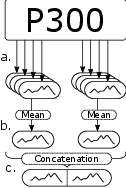}
  \caption{Scheme followed in the object definition. For each intensification of
    the two target stimuli (row-column of the target character that should
    contain a P300-ERP), our system creates a first object (A) (which will be stored in a common set). Then, all these objects were shuffled and grouped together in subsets of $M$ elements in order to average each one of them, and compose the second objects (B). Finally, the group process is repeated to concatenate each subset in groups of $C$ objects into the final objects (C). In this figure $M = 4$  and $C = 2$. This process is repeated for the non-P300 objects as well.}
  \label{fig:object-definition}
\end{figure}

Once all the labeled vectors have been stored, each object is created by
combining three operations: random 
selection of segments, average of the selected segments and concatenation of
several averages, as figure 
\ref{fig:object-definition} shows. First, several segments from one class are
randomly chosen (see figure 
\ref{fig:object-definition} (A)). Second, the selected segments are averaged (see figure
\ref{fig:object-definition} (B)). Third, several products of the averaging
process are combined to create one object (see figure
\ref{fig:object-definition} (C)). Accordingly, the object creation can be parameterized
with two parameters: number of selected segments to average ($M$) and number of
selected concatenations ($C$).

It is important to point out that the reason for applying an average process is
to improve the capabilities of extracting P300 features by reducing the inherent
noise of the signal. It 
has to be highlighted that this 
averaging approach has been widely used in other BCI solutions. On the 
other hand, the concatenation process aims to improve the accuracy of
compression distances. The idea behind 
the concatenation process is to build the object with several examples of
averages. In this way, the object 
is more general because of the different examples that it contains of the class
that it represents.

Owing to the parametrization of the 
objects which includes the number of segments used to calculate the averaged
segments ($M$), and the number 
of averaged segments concatenated ($C$), the objects may differ in each
experiment. Thus, as it will be 
described in Section \ref{sec:results}, we have explored the effect that changing these parameters has on the
quality of the obtained results.

\subsection{From the NCD to data clusters}
\label{subsec:ncd}

In this work we explore if the NCD can be used to distinguish between P300-ERPs
and non-P300-ERPs, which depends on whether compression distances are able to
capture P300 structure or not. For this purpose, we first create ASCII objects
from EEG signals, as explained in Section \ref{subsec:signalprocess}. Then, a
NCD matrix that contains 
all the pairwise distances between every pair of objects is calculated. This
matrix can be seen as a reduced form of the information representing the
original dataset, however, in this format, this information cannot be easily
analyzed. That is the reason why we use clustering to transform the information
contained in the NCD matrix to a cognitively acceptable format. In other words,
we transform a NCD matrix into data clusters. We proceed like this because the
most usual ways to measure NCD matrix grouping capabilities is through
clustering techniques such as hierarchical clustering and multidimensional
mapping \cite{cilibrasi_clustering_2005,paulovich_projection_2007}.

Therefore, in this work, we use clustering as a tool to analyze if our object
creation method suitably represents the information contained in the signals and
to explore if compression distances are a valid tool for identifying P300. In
this manner, we use the NCD matrix as input to a clustering method and we
analyze the performance of the clustering by analyzing the quality of the
clusters provided by the clustering method. In an ideal clustering all P300
objects would be clustered together, and the same would happen to all non-P300
objects. Summarizing, if our ASCII objects suitably represented the
information contained in EEG signals and the NCD were a valid tool for
identifying P300, we would obtain high quality clustering results.

With the purpose generalizing our results, we apply two different clustering
methods. First, we use the one presented in \cite{cilibrasi_clustering_2005},
developed by the creators of 
the NCD. This is a hierarchical clustering algorithm based on the minimum
quartet tree method that takes a NCD matrix as input and generates a dendrogram
as output \cite{complearn,cilibrasi_clustering_2005}. Second, we use another clustering method
that can work with 
NCD matrices. This is a multidimensional mapping algorithm based on
multidimensional projections (Nearest Neighbors Projection)
\cite{feng_projection_2016,telles_normalized_2007} that takes
a NCD matrix as input and generates a projection as output. It is important to
point out that multidimensional projections use a different approach from
hierarchical clustering methods based on dendrogram representation. Using two
clustering methods of such different nature allows us to increase the level of
generalization of our study. The following sections describes both clustering
algorithms in depth.

\subsubsection{Minimum quartet tree method (CompLearn)}
\label{subsec:maketree}

The first approach used in our work consists of using an algorithm based on the
minimum quartet tree method. In terms of implementation, we use the CompLearn 
Toolkit \cite{complearn}, which 
implements the clustering algorithm described in \cite{cilibrasi_clustering_2005}.
This algorithm has an asymptotic cost of $\mathcal{O}(N^3)$ (CompLearn version
1.1.5). Therefore, there is a limitation in the size of the data sets, due to
the convergence costs of this algorithm. Once the NCD matrix is calculated, it is
used as input to the clustering phase and a dendrogram is generated as output.

\begin{figure}
  \centering
  \includegraphics[width=0.7\textwidth,trim={0 1cm 0 0},clip]{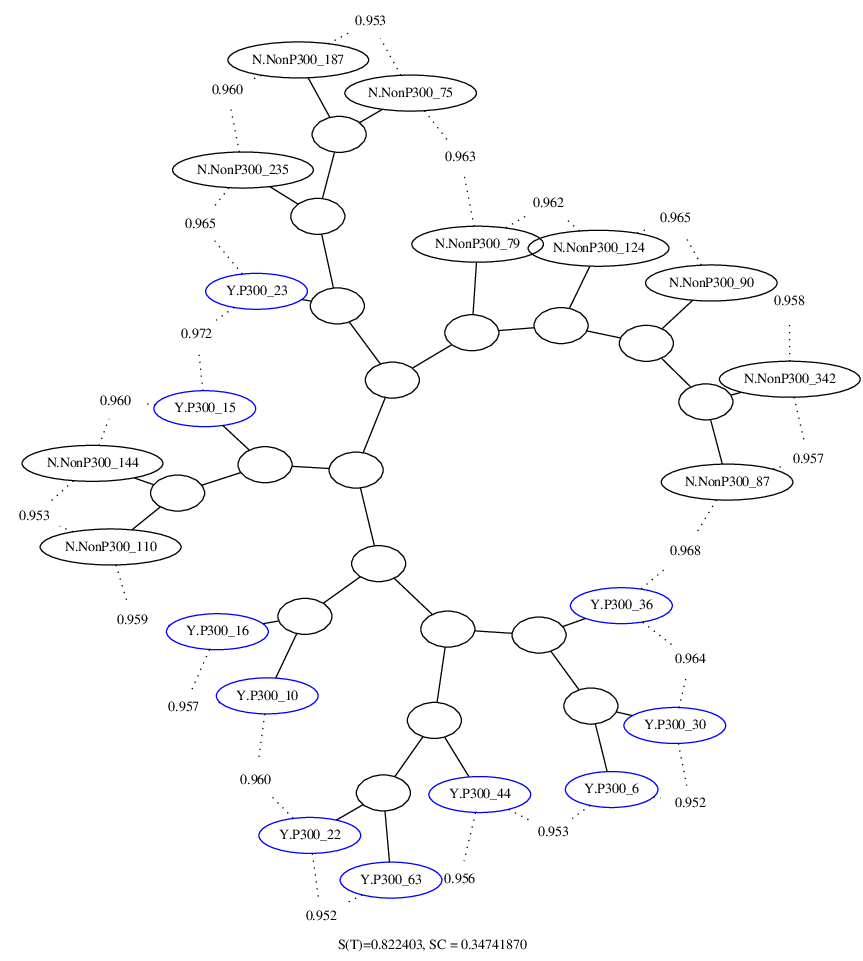}
  \caption{Sample of a CompLearn's hierarchical tree \cite{complearn} from a subset of objects, of the second BCI Competition. CompLearn is a
    compression based toolbox that takes, in this case, the NCD distance matrix as
    input and generates a hierarchical tree. The configuration used was $C = M =
    4$, according to the method described in figure \ref{fig:object-definition}.
    The blue nodes correspond to the P300s objects and the black ones to the
    non-P300. The Silhouette Coefficient of the dendrogram is 0.3474. } 
  \label{fig:maketree}
\end{figure}

A dendrogram is an undirected binary tree diagram, frequently used for hierarchical clustering, that
illustrates the arrangement of the clusters produced by a clustering algorithm.
In figure \ref{fig:maketree}, we can observe an example of dendrogram obtained
from the BCI Competition II data set \cite{bci_ii}, using an object
configuration of C = M = 4 (a combination of four concatenations and four
averages of ERP segments for each object). That is, each
object is  
created by averaging 4 segments to obtain the ``averaged segments'' and
concatenating 4 ``averaged segments''  afterwards. 
In our work, the ``quality'' of any dendrogram obtained as output from this
process, has been measured using an adaptation of the Silhouette
coefficient (taking the path between leaves as their distance), as explained in Section \ref{subsec:silcoef}. 

\subsubsection{Multidimensional projections (PEx)}
\label{subsubsec:pex}

The second approach consists of visualizing high-dimensional data through
mapping techniques taking the
NCD matrix as input to build a multidimensional projection.
Several ways of visualizing high-dimensional data 
through mapping techniques exist in the literature. In general
terms, a mapping technique gives each datapoint a location in a two or three
dimensional map \cite{Paulovich07}. This allows
a human to visually analyze datasets because they are represented in a two or
three dimensional map. In terms of implementation, we use the Projection Explorer
(PEx), a visualization tool that takes a distance matrix as input and generates a projection through mapping
techniques \cite{Paulovich07,Telles07}.

\begin{figure}
  \centering
  \includegraphics[width=0.6\textwidth,trim={0cm 1cm 12cm 0.64cm},clip]{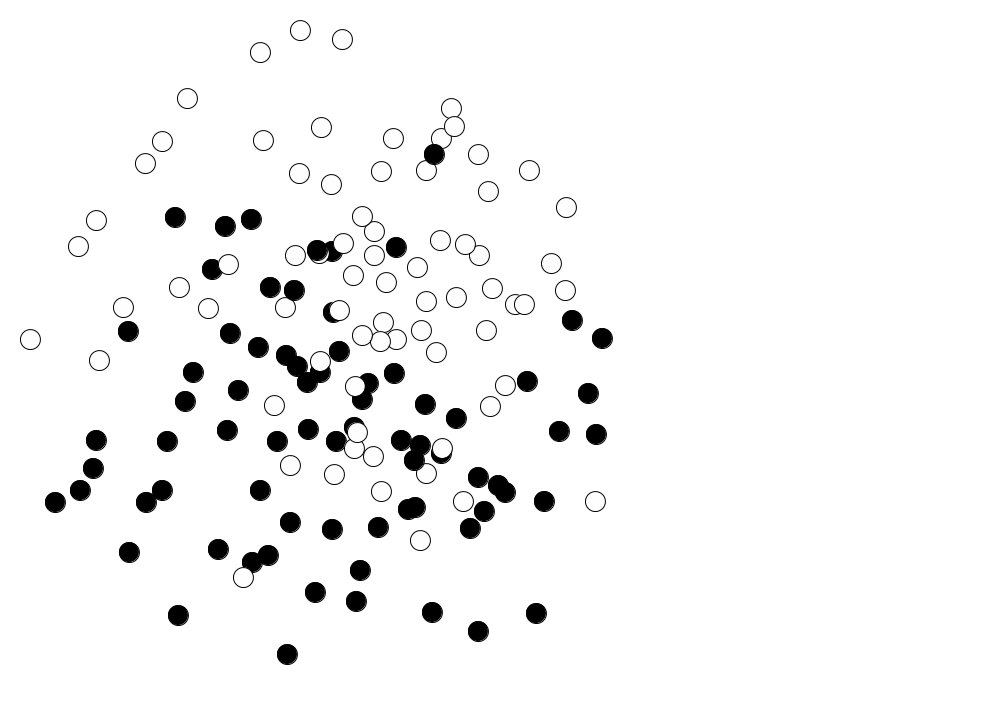}
  \caption{Sample of a PEx's projection from a subset of objects, of the second BCI Competition. PEx is a
    visualization tool that takes, in this case, the NCD distance matrix as
    input and generates a projection through mapping techniques.
    \cite{telles_normalized_2007,paulovich_projection_2007}.
    Following the
    method described in figure \ref{fig:object-definition}, we take the same
    object configuration of figure \ref{fig:maketree}, where $M = C = 4$. The
    white nodes correspond to the P300s objects and the black ones to the
    non-P300s. The Silhouette Coefficient of the projection is 0.1352}
  \label{fig:pexsample}
\end{figure}

This approach allows us not only to corroborate that the NCD can be used to
detect P300 in EEG but also to 
represent more samples in a visually resilient way. As mentioned above, PEx, in
contrast to CompLearn, has fewer limitations about the number of objects 
used in the analysis. In figure \ref{fig:pexsample}, we show an output example
of PEx over a subset of objects previously built as 
described in Section \ref{subsec:signalprocess}. The configuration of the
objects is $C = M = 4$, and like in figure \ref{fig:maketree}, it represents an
preliminary result of our work. In this case the Silhouette
Coefficient is measured by the PEx software using the Euclidean distance
between each pair of objects. 

\subsection{Data cluster quality in the Dendrogram: Silhouette Coefficient}
\label{subsec:silcoef}

The main objective of this work is to explore whether a compression-based
clustering methodology could extract a single P300-ERP structure regardless of
the variability contained in the EEG signal. Then, using the described methods
in the previous section, this can only happend if the output of the clustering
shows a single cluster for the P300 group. Thereby, we make use of the
Silhouette Coefficient (SC)  \cite{rousseeuw_silhouettes:_1987}
to measure the cohesiveness of each
cluster compared to the other cluster. Thus, the coefficient which defines the relation
of an object with the clusters is defined as follows:

% \%%vspace{-0.2cm}
\begin{eqnarray}
  s(i) = \frac{b(i) - a(i)}{max\{a(i),b(i)\}}, 
\end{eqnarray}
% \%%vspace{0.2cm}

\noindent where the measure $a(i)$ is the average distance of $i$ with all other
objects within the same cluster, and  $b(i)$ is the average lowest distance of
$i$ to the nearest other cluster. Finally, the
Silhouette coefficient of the whole dataset is calculated averaging $s(i)$ over
all objects of the entire dataset.

Hence, the score given by the SC will
be reinforced by low values of $a(i)$ and high
values of $b(i)$. In other words, by similar objects
within the cluster and dissimilar clusters within the set.
The PEx toolkit \cite{paulovich_projection_2007} uses the average Euclidean
distance between nodes (objects in 
this case) to calculate $a(i)$ and $b(i)$.  

However, in order to measure the SC from a dendrogram we have
adapted its general definition to our specific domain, following
\cite{granados_discovering_2015}. Thus, with the aim of calculating the SC in a dendrogram, we
define the distance, $d\{i,j\}$, between each 
pair of objects (or dendrogram leaves), as the
number of nodes between $i$ and $j$. Thus, the distances $a(i)$ and $b(i)$ have
been defined as follows for this specific case: 

% \%%vspace{-0.2cm}
\begin{eqnarray}
~~a(i) = \frac{1}{n} \sum\limits_{j~\in~C_i}{d\{i,j\}} ~,~~ b(i) = \frac{1}{m}
\sum\limits_{j~\notin~C_i}{d\{i,j\}}~,\nonumber
\end{eqnarray}
% \%%vspace{-0.2cm}

\noindent where $n$ is the number of elements of its cluster ($C_i$) and $m$ is
the number of elements of the cluster more similar to object $i$ and different
from $C_i$.

\section{Experimental results}
\label{sec:results}

In this section, we show the experiments that have been carried out.
First, we explore how changes in the number of segments used to assemble the objects
affect the identification of P300 in the EEG signal. For this analysis a
single electrode as data source is used. Second, we 
study the activity of P300 events in all the 64 electrodes used in standard
BCI schemes. Third, we combine the information of several electrodes to study
the optimal objects' configuration. That is, we perform a detailed study on the
effects of changing the parameters M and C when several electrodes are used.
Finally, we carry out some experiments 
over a different dataset to generalize our results. 

\subsection{Preliminary study: means and concatenations}
\label{subsec:sys_test}

The purpose of this preliminary study is to gain intuition about how much
information can be identified in the objects 
depending on the parameter configuration used to create them. That is, depending
on the values of $M$ (means) and $C$ (concatenations) used to create the objects
as described in Section 
\ref{subsec:signalprocess}. In order to do this, we have briefly studied the
impact that changing the parameters $M$ and $C$ has on the clustering quality. In
this regard, two different experiments have been carried out. %First, we have produced four sets of objects

The objects for both experiments have been generated using a single electrode as
data source. More 
particularly, we have taken a reference data source for the detection of the
P300, the so-called Cz electrode position according
to \cite{lagerlund_determination_1993}. We have chosen this electrode because it
is well known that it reports good results in the classification of P300-ERPs
\cite{krusienski_critical_2011,didone_auditory_2016}. 

After creating the objects, we have
calculated the NCD for every pair of nodes. Then, using this matrix as input, we have
clustered the objects
by means of two different methods: CompLearn and PEx, obtaining dendrograms
and projections, respectively, as output to the methods. Finally, the
clustering quality has been quantified using the 
Silhouette 
Coefficient (SC). 

\begin{figure}
  \centering
  \begin{tabular}{|c|c|}
    \hline
    \includegraphics[width=0.25\textwidth,trim={0cm 0cm 12cm 1.25cm},clip]{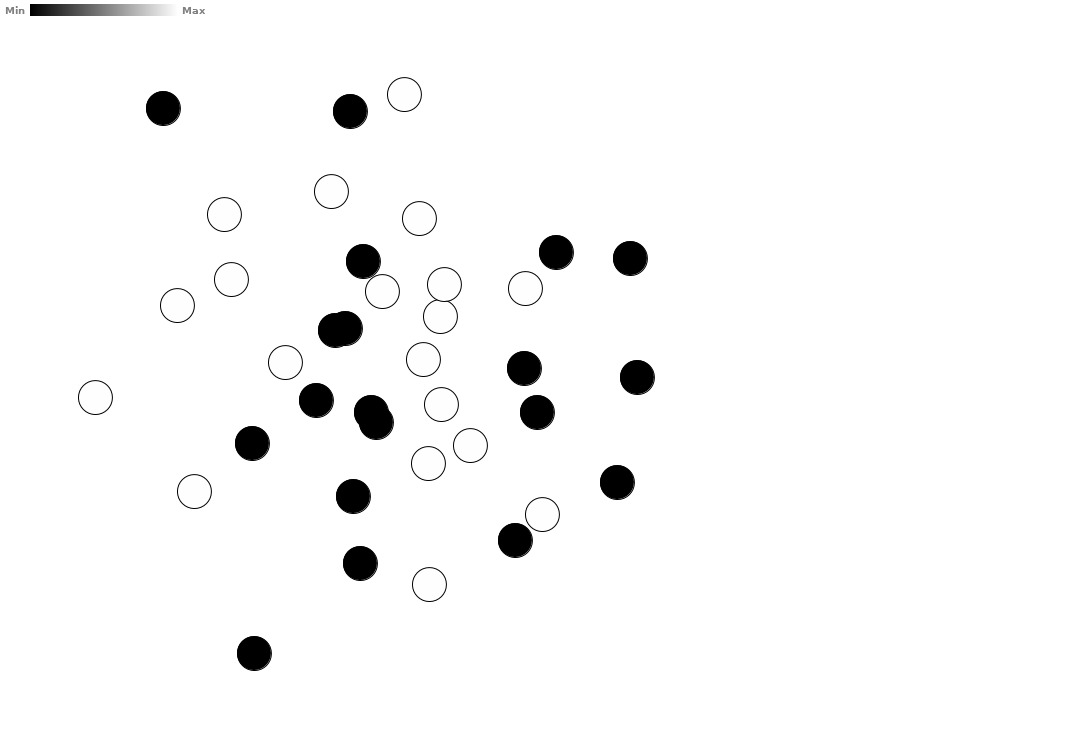}&
                                                                                                     \includegraphics[width=0.25\textwidth,trim={ 0cm 0cm 10cm 1.cm},clip]{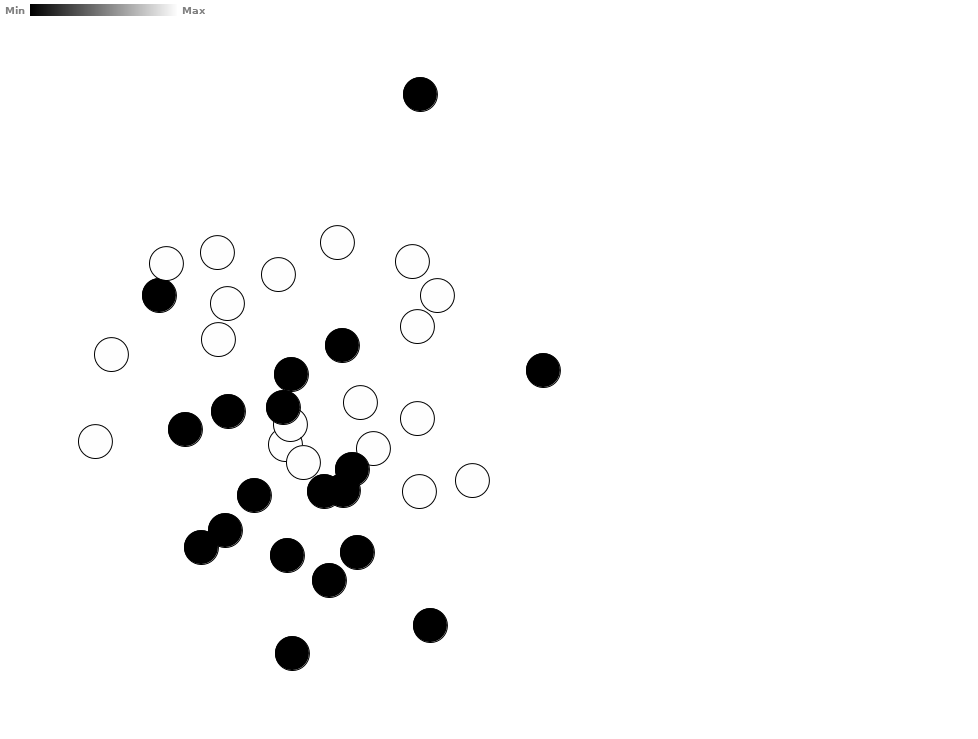}\\
    $C = M = 1$ & $C = 1 ; M = 8$ \\
    \hline
    \includegraphics[width=0.25\textwidth,trim={ 0cm 0cm 13cm 1.25cm},clip]{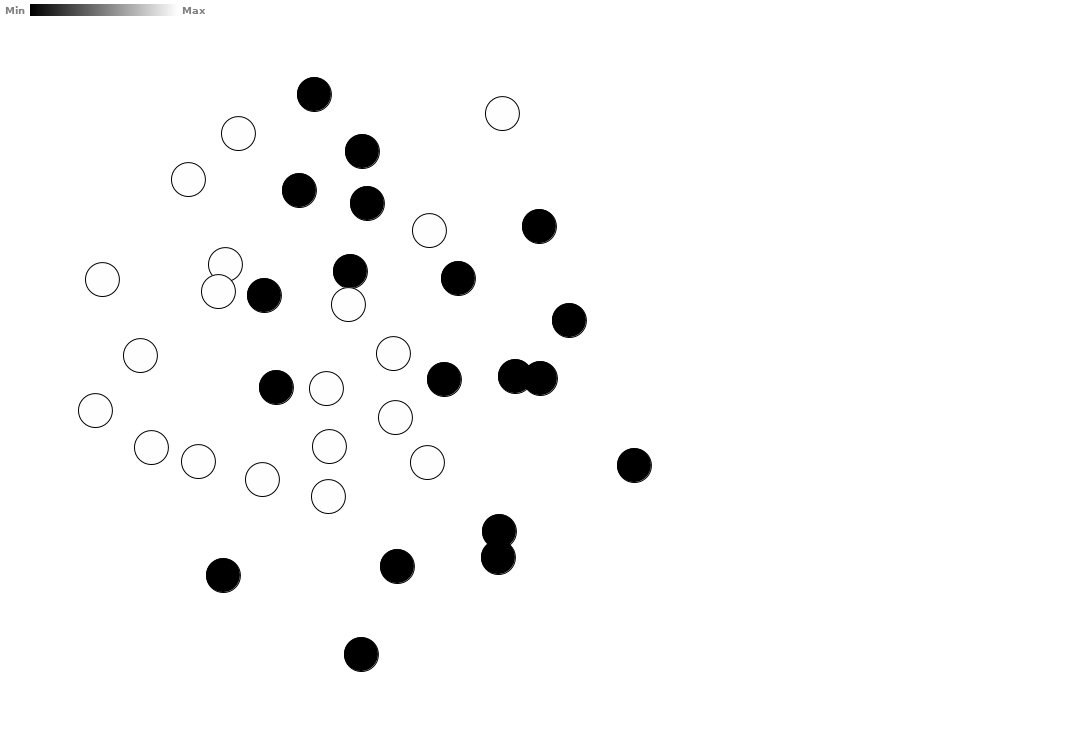}&
                                                                                                         \includegraphics[width=0.25\textwidth,trim={ 0cm 0cm 14cm 2.5cm},clip]{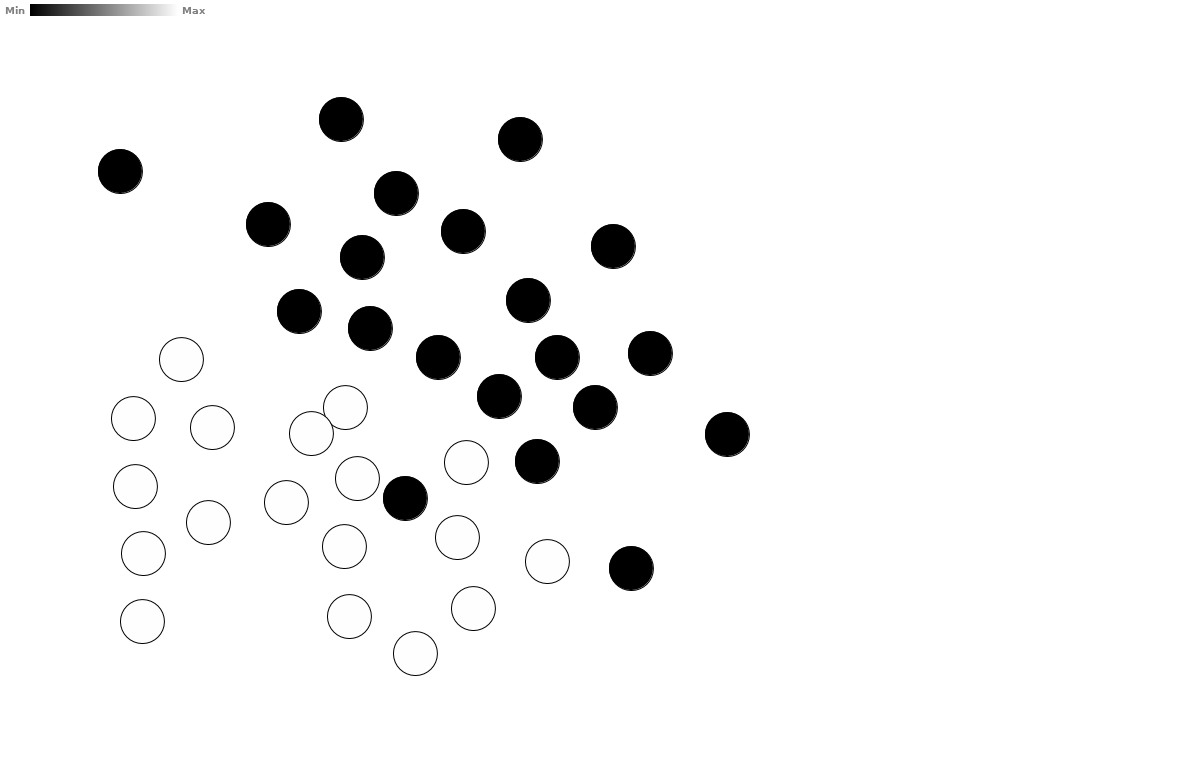}\\
    $C = 8 ; M = 1$ & $C = M = 8$ \\
    \hline
  \end{tabular}
  \caption{Output of PEx from different object configurations, for a single
    electrode
    (Cz according to \cite{lagerlund_determination_1993}). We observed
    in this experiment that the increment of segments per object seems to
    improve the separation of both clusters. In order to compare each
    configuration we take a subset of objects for each one of them due to the
    last configuration, Lower-right, that can not generate more than 18 P300 objects. }
  \label{fig:pex_comparison}
\end{figure}

For the first experiment, and due to the limited 
number of P300-ERPs in the dataset, we limited the number of segments per object
to 64 to maintain a minimum number of objects per projection. Thus, we
have created objects using four different configurations: (i) $M = C = 1$, (ii)
$M = 1, C = 8$, (iii) $M = 8, C = 1$, (iv) $M = C = 8$. By visual inspection
of figure \ref{fig:pex_comparison}, it can be
observed that incrementing the number of segments used to create the objects,
seems to improve the separation of P300 and non-P300 objects. In this case, the
configuration which gives us the best projection is the one with $M = C = 8$.
This phenomenon is also slightly visible in the raw 
distances between P300 and non-P300 objects.

\begin{table}
  \centering
  \begin{tabular}{ | c| c| c|| c|}
    \hline
  \textbf{C = M = 1}   & \textbf{P300s} & \textbf{non-P300s} & \textbf{diff} \\ %
    \hline
    \textbf{P300s} & 0.98612 & 0.98617 & $\sim 10^{-5}$  \\
    \hline
    \hline
  \textbf{C = M = 1}   & \textbf{P300s} & \textbf{non-P300s} & \textbf{diff} \\ %
    \hline
    \textbf{P300s}& 0.95246 & 0.95592  & $\sim 10^{-3}$ \\ %
    \hline
  \end{tabular}
  \caption{Average NCD between the different pairs of groups of objects for two different
    configurations. The upper table shows the
    different values measured in a sample of objects assembled with C = M = 1
    configurations. In this case one can 
    observe a small difference between the average inner-group distance and
    inter-group distance ($\sim 10^{-5}$). On the other
    hand, using a higher configuration (C = M 
    = 8), depicted in the lower table, the differences between these two
    measures are bigger ($\sim 10^{-3}$). Thus, looking at the average difference distance
    between groups (forth column), we can see an improvement of 100 times in the
    second case compared with the first one. An example of these two $C$ and $M$
    configurations is 
    depicted in figure \ref{fig:pex_comparison}.}
  \label{table:averdistance}
\end{table}

Looking at table
\ref{table:averdistance}, one can notice that the average NCD between the
two groups is higher as the number of ERPs per object grows. Particularly, in
the first case ($C = M = 1$) the difference is $\sim 10^{-5}$ while in the
second case ($C = M = 8$) is
$\sim 10^{-3}$. This means that the objects are 100 times different (on average) in the
second case compared with the first one. Which means that the resolution of the
NCD will be better and thereby, the final clustering.

\begin{figure}[h!]
  \centering
  \includegraphics[width=0.78\textwidth,trim={0.5cm 0.cm 0.8cm 0.5cm },clip]{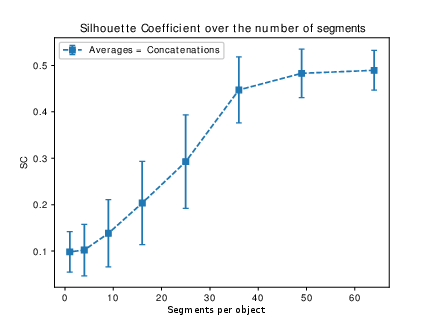}
  \caption{Simulation over the number of segments per object (single P300-ERPs
    or non-P300-ERPs) for an unique electrode
    (Cz according to \cite{lagerlund_determination_1993})
    through Maketree's clustering. Each data point represents the average of
    iterating 100 times the simulation (20 files per simulation due to cost
    limitations of Maketree algorithm). The data was taken
    from both training sessions (10 and 11). The iterated parameters were the
    number of segments used in both concatenation and average processes. In this
    case, the curve only represents those configurations with equal number of
    segments for both concatenation and average processes (i.e. $M = C$).}
  \label{fig:dot}
\end{figure}

In order to dig into this preliminary result, we have created more sets of objects using more configurations,
with $M = C$, and we have studied how the clustering quality depends on the number of segments used to create
the objects. We have explored only configurations where $M = C$ because the purpose of this preliminary study
is not to analyze the impact of $M$ and $C$ separately, but to corroborate that incrementing the number of
segments used to create the objects improves the quality of the clustering
results (figure \ref{fig:dot}). It can be observed that the
clustering quality improves as the number of segments increases, as the first experiment suggested. However,
it is necessary to carry out a broad study on the effects of changing the
parameters $M$ an $C$ separately using more 
electrodes to derive more conclusions (see Section \ref{subsec:object_comp}).

We have to point out that for the analysis presented in this section, we have
used  the well-known Cz electrode because the P300-ERP is measured very strongly
in this electrode. A logical question that arises is: what happens with other
electrodes where P300-ERP is not measured so strongly. The goal of next section
is to analyze this question by extending the previous analysis to all the 64
electrodes to observe if the differences in the strength of P300-ERP measurement
are congruent with the quality of the clustering obtained. 

\subsection{Scoring sources}
\label{subsec:source_sim}

The purpose of this experiment is studying the activity of the P300 in all the
electrodes. The experiment 
that has been carried out is similar to the last experiment presented in
previous section. 
That is, the parameters M and C are equal, particularly, $M = C = 8$.

For each electrode, we have limited the number of objects per dendrogram to 20.
In this case, the number of objects have been chosen to
overcome the convergence limitations of CompLearn.
Thus, as a consequence of this, we have repeated the process 100 times.
In order to summarize the results of each electrode, we have measured the SC of
each dendrogram and depicted each distribution both as a boxplot and a scalp
colormap. This allows an easy comparison of the electrodes depending on their
position on the scalp. 

\begin{figure}[h!]
  \centering
  \begin{tabular}{c}
     \includegraphics[width=0.9\textwidth,trim={ 0cm 0cm -.8cm 0cm},clip]{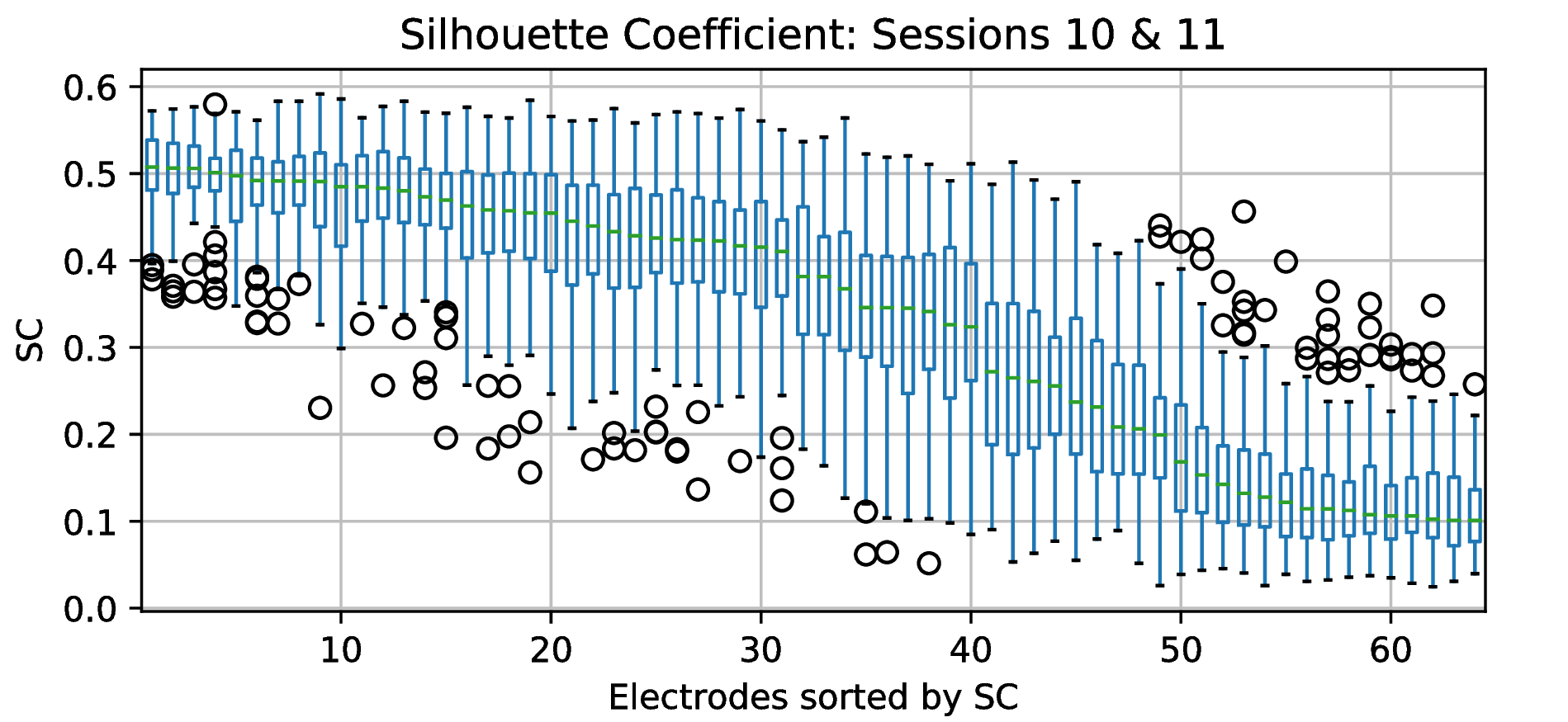}\\
    \includegraphics[width=0.75\textwidth,trim={ .1cm 0.cm -0.5cm 0cm},clip]{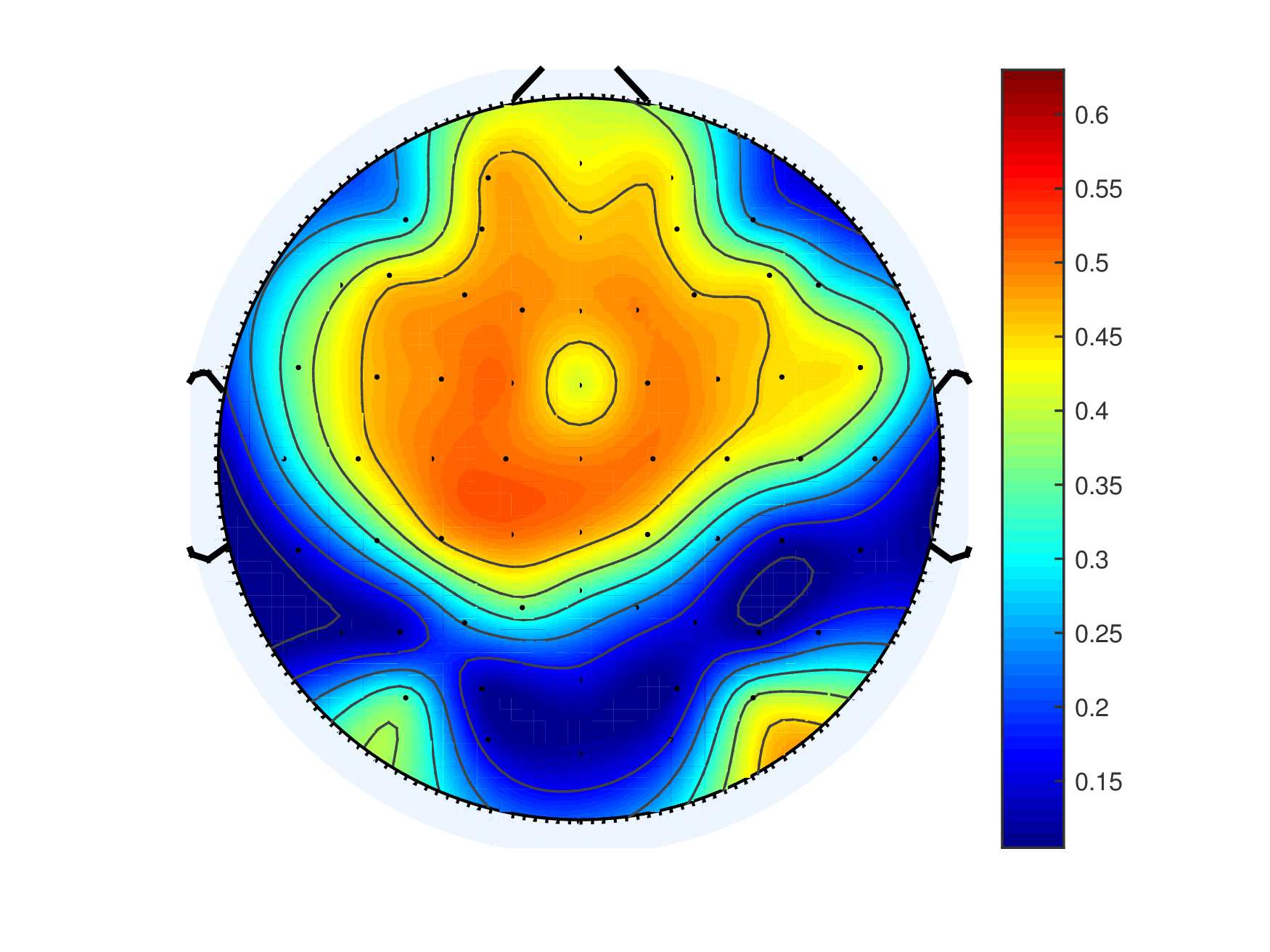}\\
  \end{tabular}
  \caption{Silhouette Coefficient distribution obtained for each electrode for
    both sessions 10 and 11. Each score was calculated from a $M = C = 8$
    configuration from 100 iterations of CompLearn simulation. The upper figure
    shows the SC distribution among electrode through boxplots. The lower figure
    shows the median of the SC for each electrode across the scalp.}
  \label{fig:electrodesim}
\end{figure}

Figure \ref{fig:electrodesim} depicts the SC obtained for all the 64 electrodes,
using the data of both 
sessions (see Section \ref{subsec:dataset_met}). Looking at both the boxplot and scalp distributions, a couple of
observations can be made. 
First, due to the SC distribution, we notice that the structure of the P300-ERP is not
manifested equally among the electrodes, as it is well known
\cite{blankertz_single-trial_2011,mccann_electrode_2015,krusienski_critical_2011}.
Second, we observed that there are three main areas (corresponding to the
warm colors in the scalp) where the structure of the P300-ERP appears more
clearly. These areas are similar to the ones obtained in other works of P300-EEG
context 
\cite{mccann_electrode_2015,li_age-related_2013}.
This fact, supports the idea of a good ERP structure identification by our
methodology. This phenomenon also appears in the individual analysis for
sessions 10 and 11.

\begin{figure}
  \centering
  \begin{tabular}{c}
    \includegraphics[width=0.88\textwidth,trim={ .9cm 0.cm .8cm 0.cm},clip]{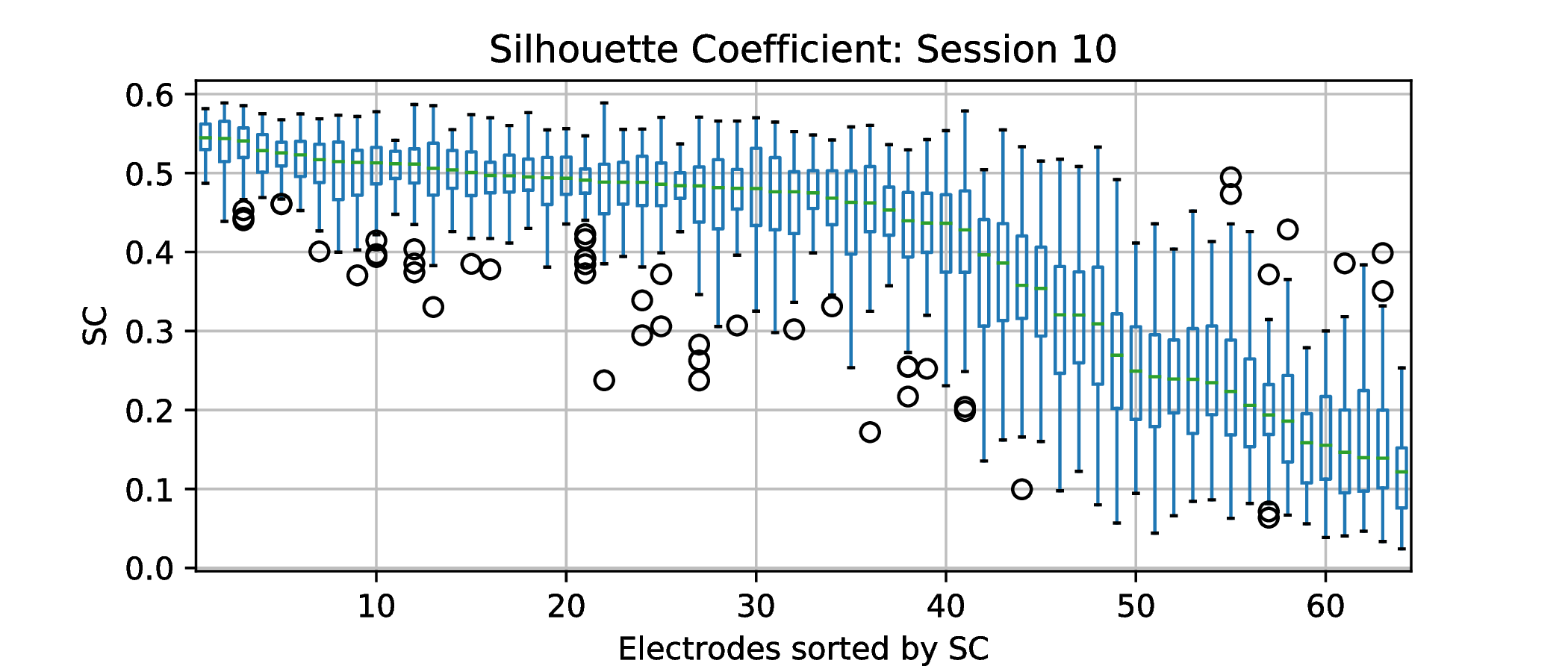}\\
    \includegraphics[width=0.88\textwidth,trim={ 0.9cm 0cm .8cm 0cm},clip]{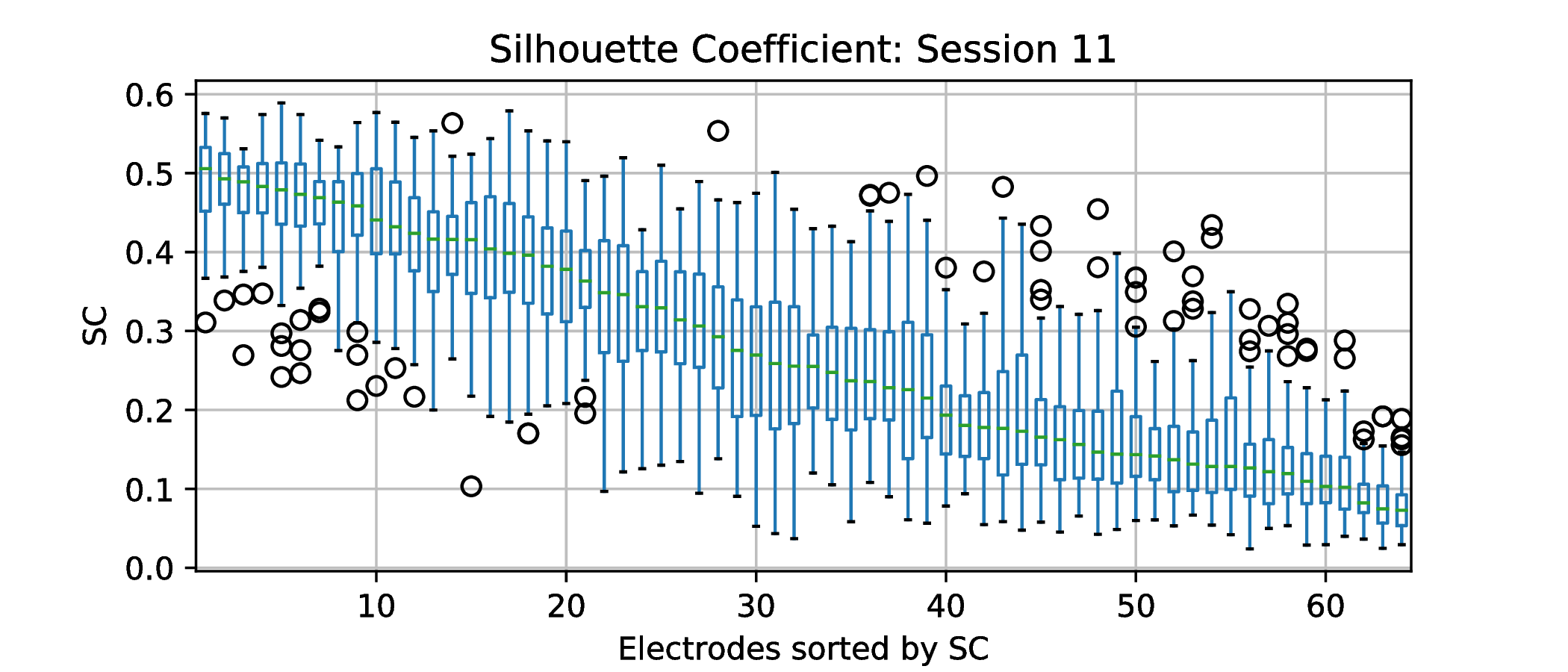}\\
  \end{tabular}
  \caption{Silhouette Coefficient boxplot distribution obtained for each electrode for sessions 10 and 11, individually. Each score was calculated from a $C = M = 8$ configuration from 100 iterations of CompLearn simulation.}
  \label{fig:electrodesim1110}
\end{figure}

\begin{figure}
  \centering
  \begin{tabular}{cc}
  \includegraphics[width=0.4725\textwidth]{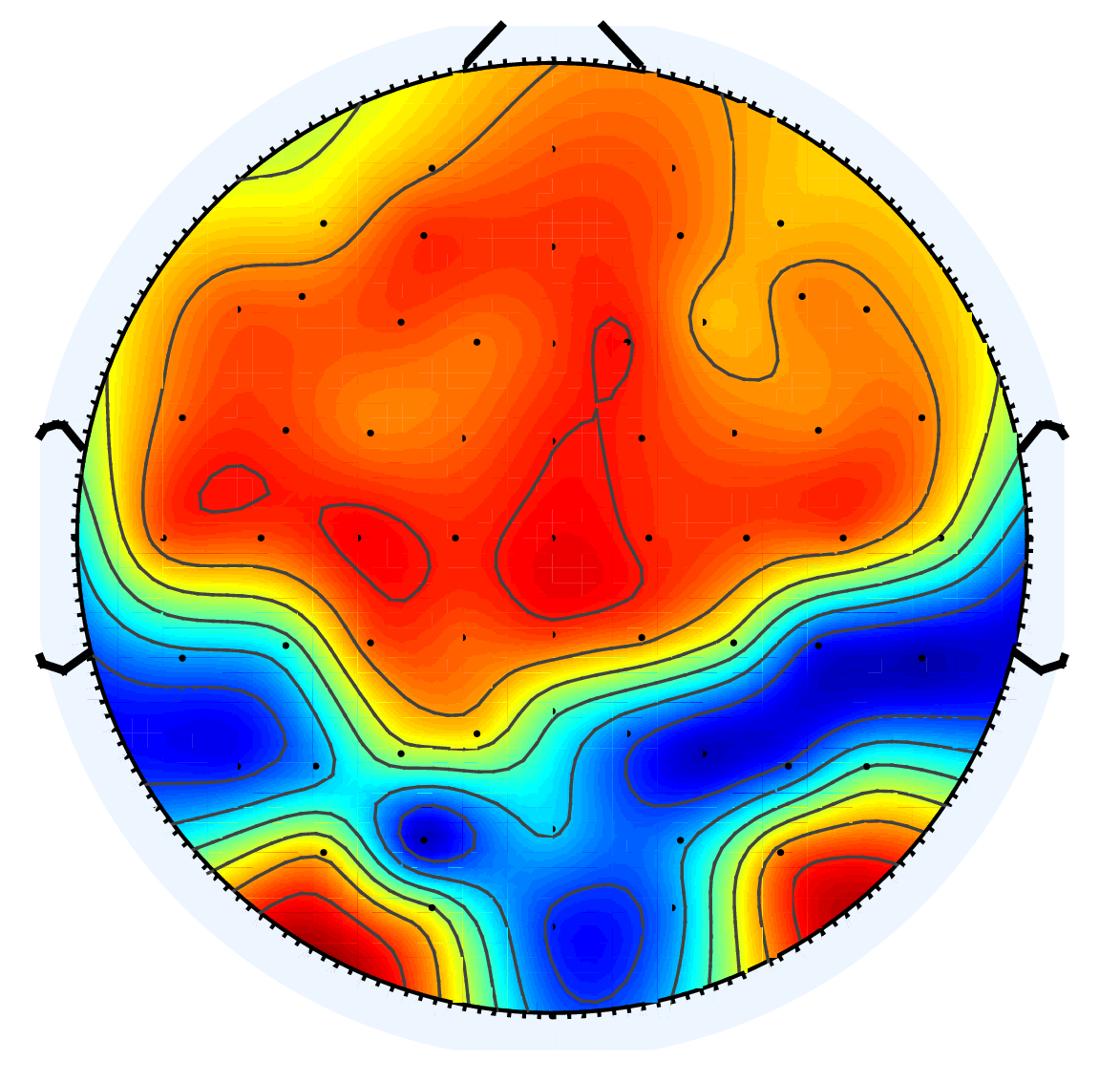}& %,trim={ 3cm 0cm 2cm 0.5cm},clip
  \includegraphics[width=0.555\textwidth]{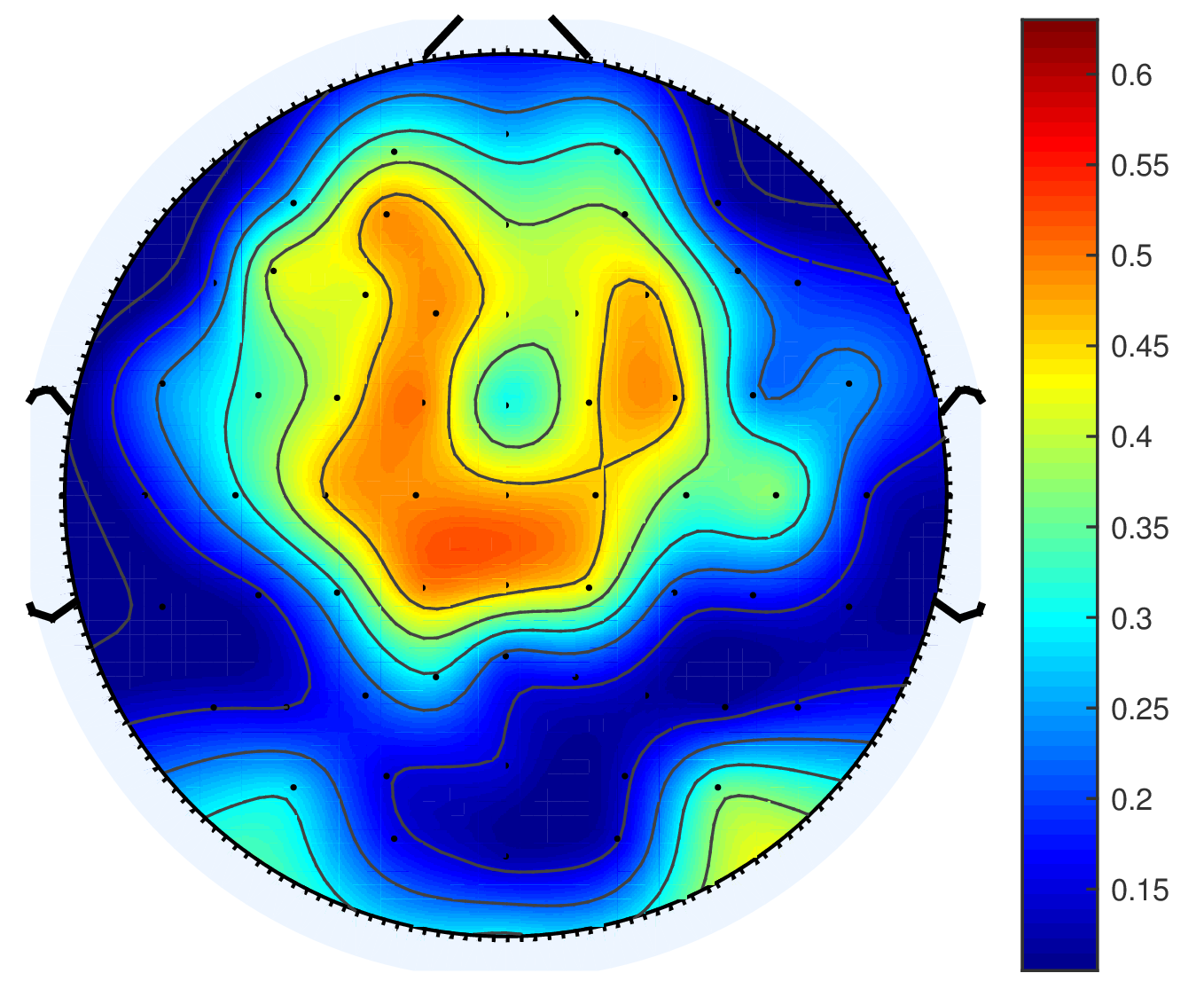}\\ %,trim={ 2.1cm 0cm 0cm 0.5cm},clip
  \end{tabular}
  \caption{Silhouette Coefficient means distribution across the scalp obtained for each electrode for sessions 10 (left) and 11 (right), individually. Each score was calculated from a $C = M = 8$ configuration from 100 iterations of CompLearn simulation.}
  \label{fig:electrodesim1110topo}
\end{figure}

Analyzing each session alone, as it is shown in figures
\ref{fig:electrodesim1110} and \ref{fig:electrodesim1110topo}, one can observe
that the 
objects created from the data corresponding to session 10 provide better
clustering results than the data from session 11 according to
\cite{xu_bci_2004}. This difference between sessions may be caused by several
reasons. Among many others, there may be a 
relocation of the electrodes or some changes in the mental state of the
subject. In this case, giving these results, we believe
that, in general, the session 11 may contain a worse signal quality
than the first one. 


Summarizing, in this section we have explored how the P300 signal is distributed along all
electrodes for a fixed configuration of objects ($M = C = 8$). Hence, for the
single-electrode analysis, we have observed that the P300-ERP is not
 equally distributed among the electrodes, as it is well know. This fact can be
 seen in figures \ref{fig:electrodesim}, \ref{fig:electrodesim1110} and \ref{fig:electrodesim1110topo}.
 Also, we have located some areas where the ERP activity is clearer, which agree
 with previous results reported in the literature
\cite{mccann_electrode_2015,li_age-related_2013}. In the next
section we will study which combination of concatenations 
and means, in the creation of objects, improves the separation between P300 and
non-P300.

\subsection{Grid Search in Object Comparison}
\label{subsec:object_comp}

\begin{figure}
  \centering
  \includegraphics[width=0.7\textwidth,trim={1.cm .5cm 2.7cm 0.cm},clip]{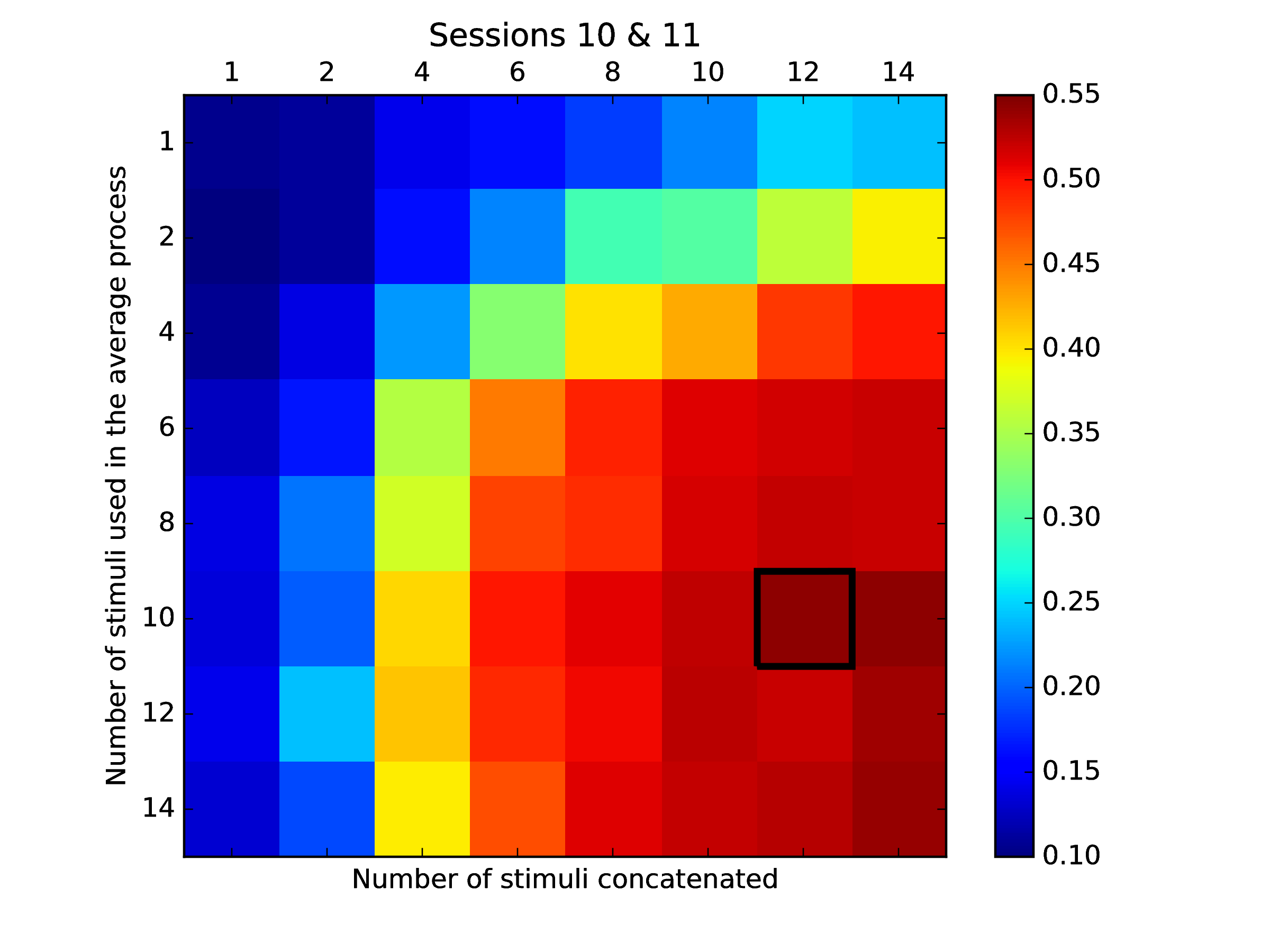}
  \begin{tabular}{ccc}
  \includegraphics[width=0.2\textwidth,trim={430px 50px 430px 50px},clip]{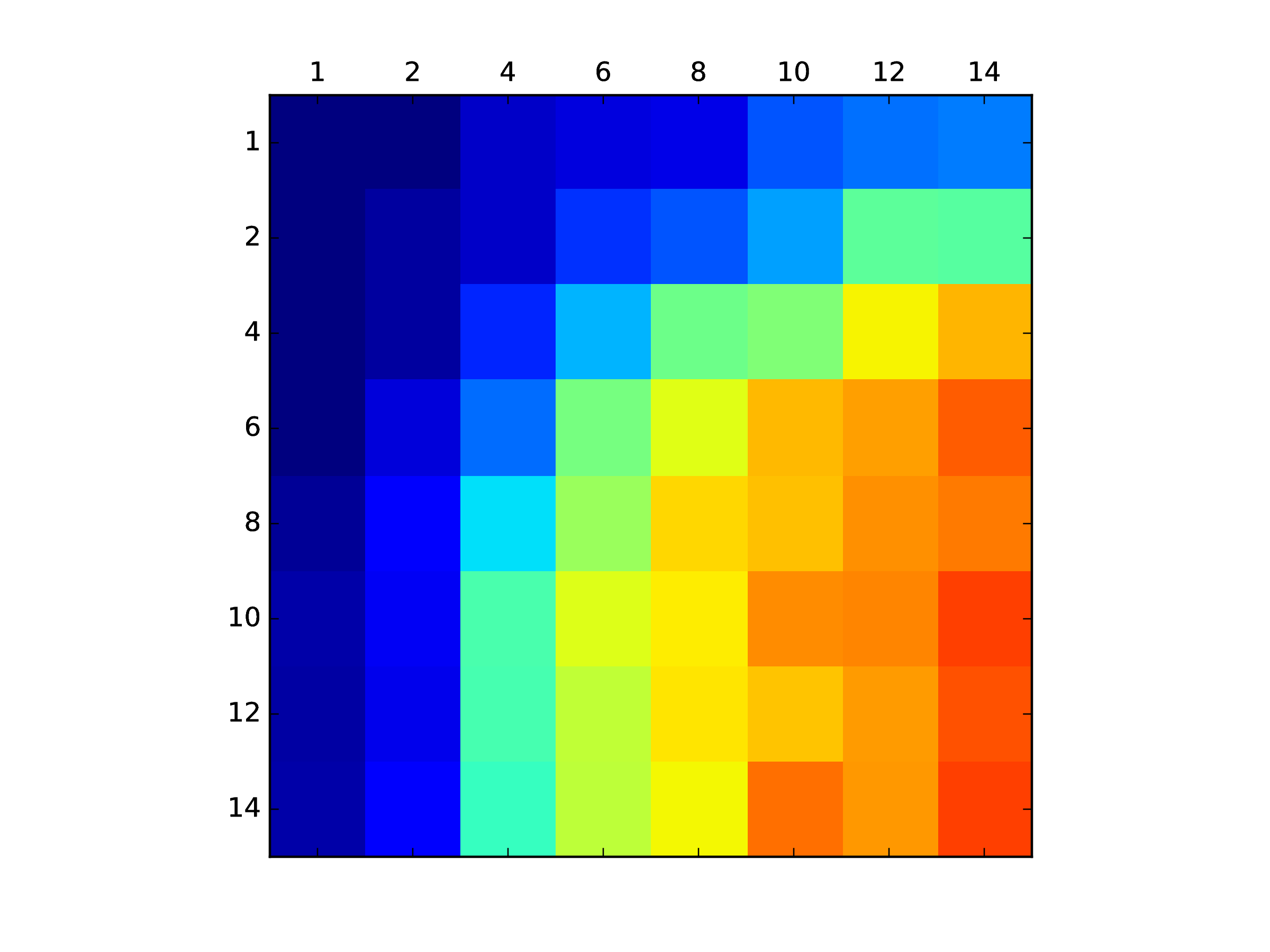}&
  \includegraphics[width=0.2\textwidth,trim={430px 50px 430px 50px},clip]{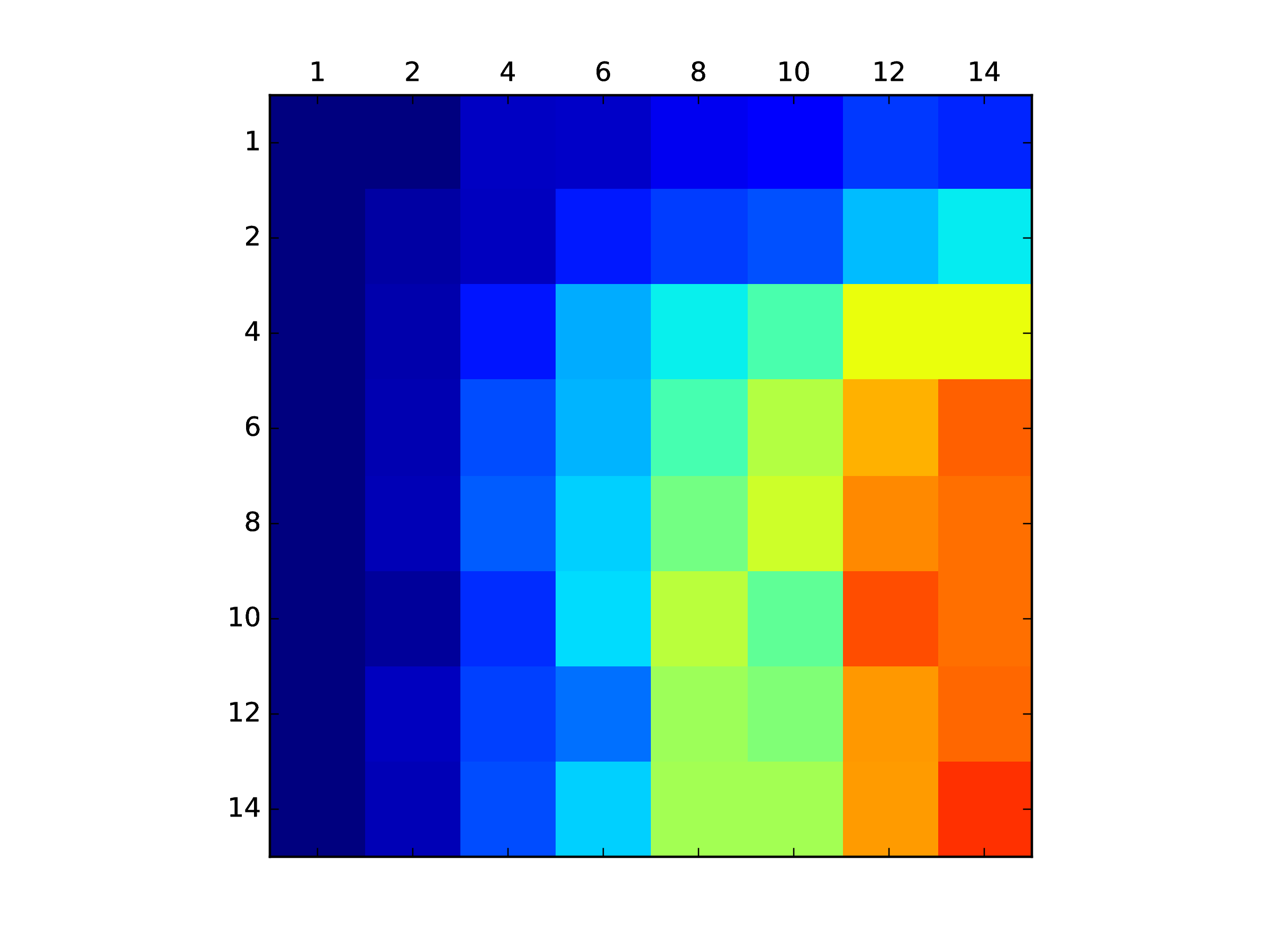}&
  \includegraphics[width=0.2\textwidth,trim={430px 50px 430px 50px},clip]{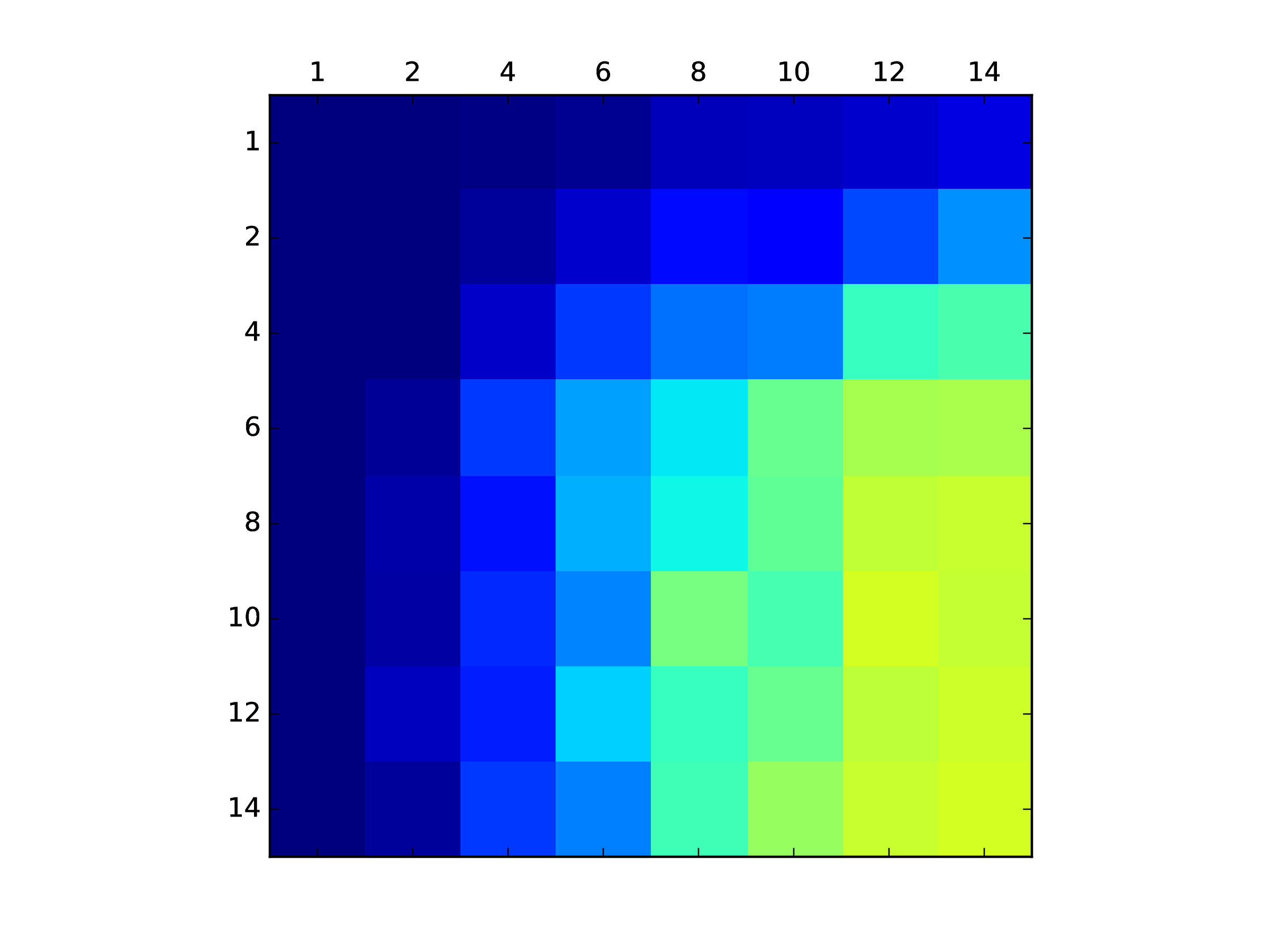}\\
  \includegraphics[width=0.2\textwidth,trim={430px 50px 430px 50px},clip]{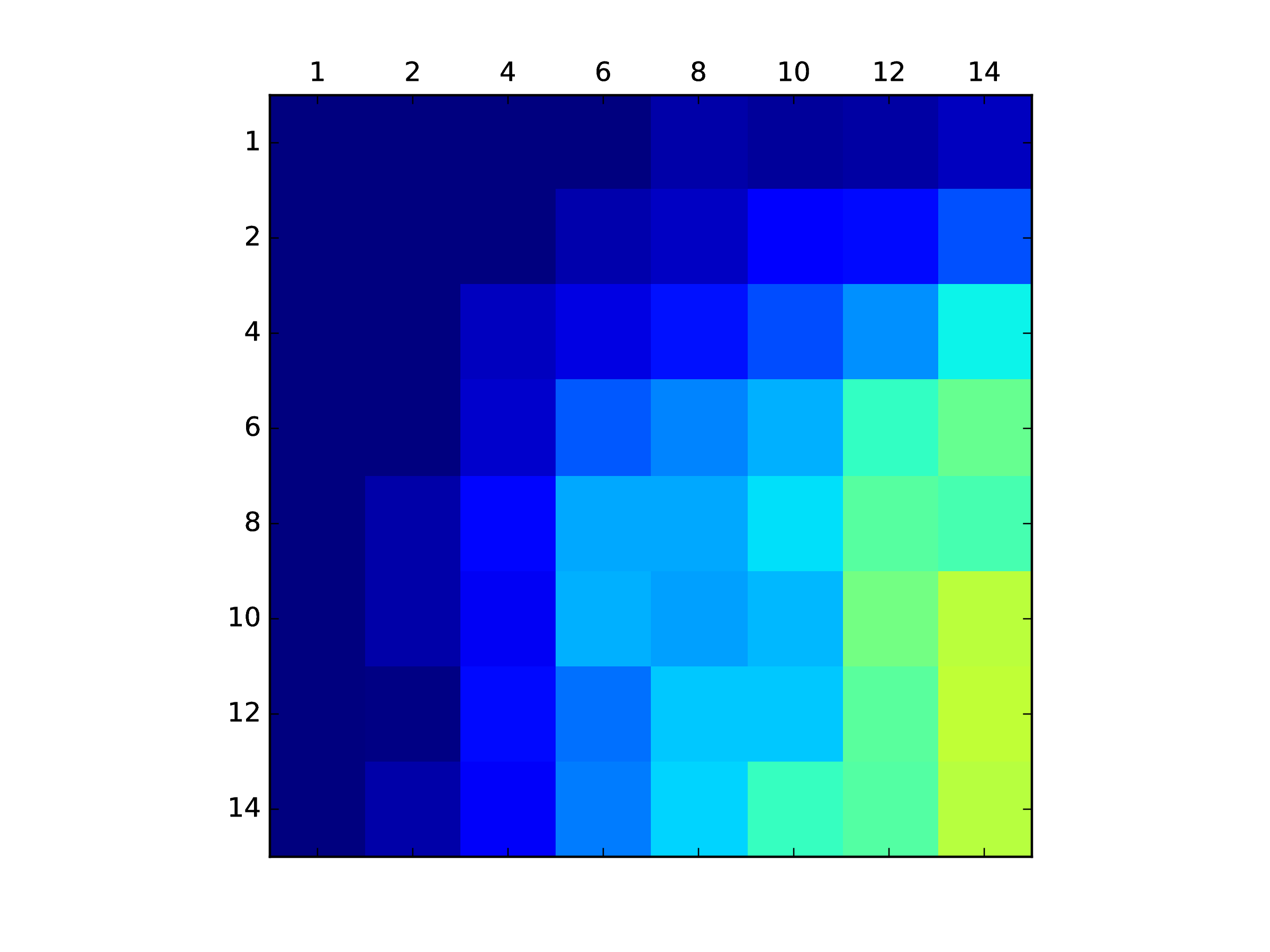}&
  \includegraphics[width=0.2\textwidth,trim={430px 50px 430px 50px},clip]{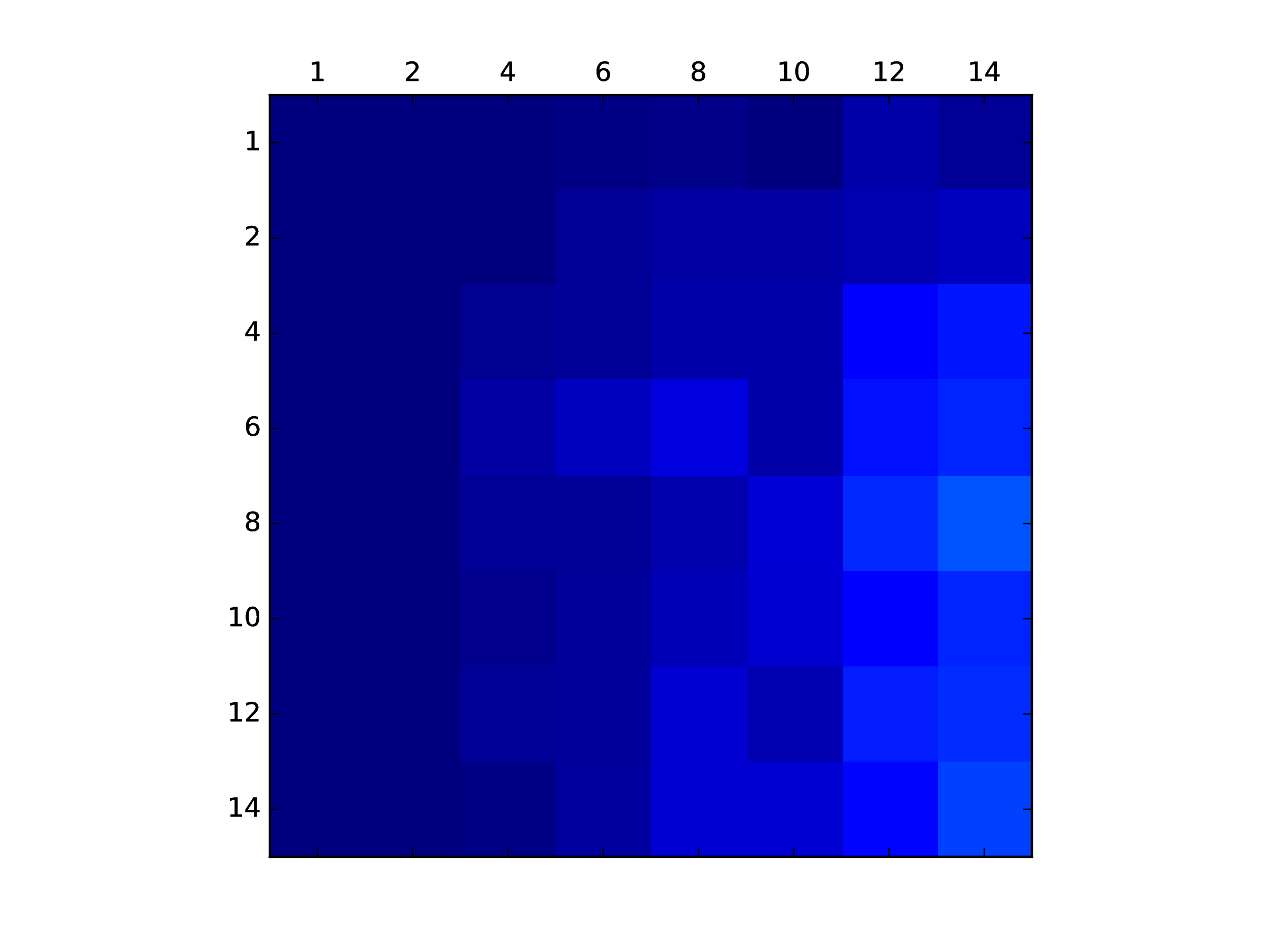}&
  \includegraphics[width=0.2\textwidth,trim={430px 50px 430px 50px},clip]{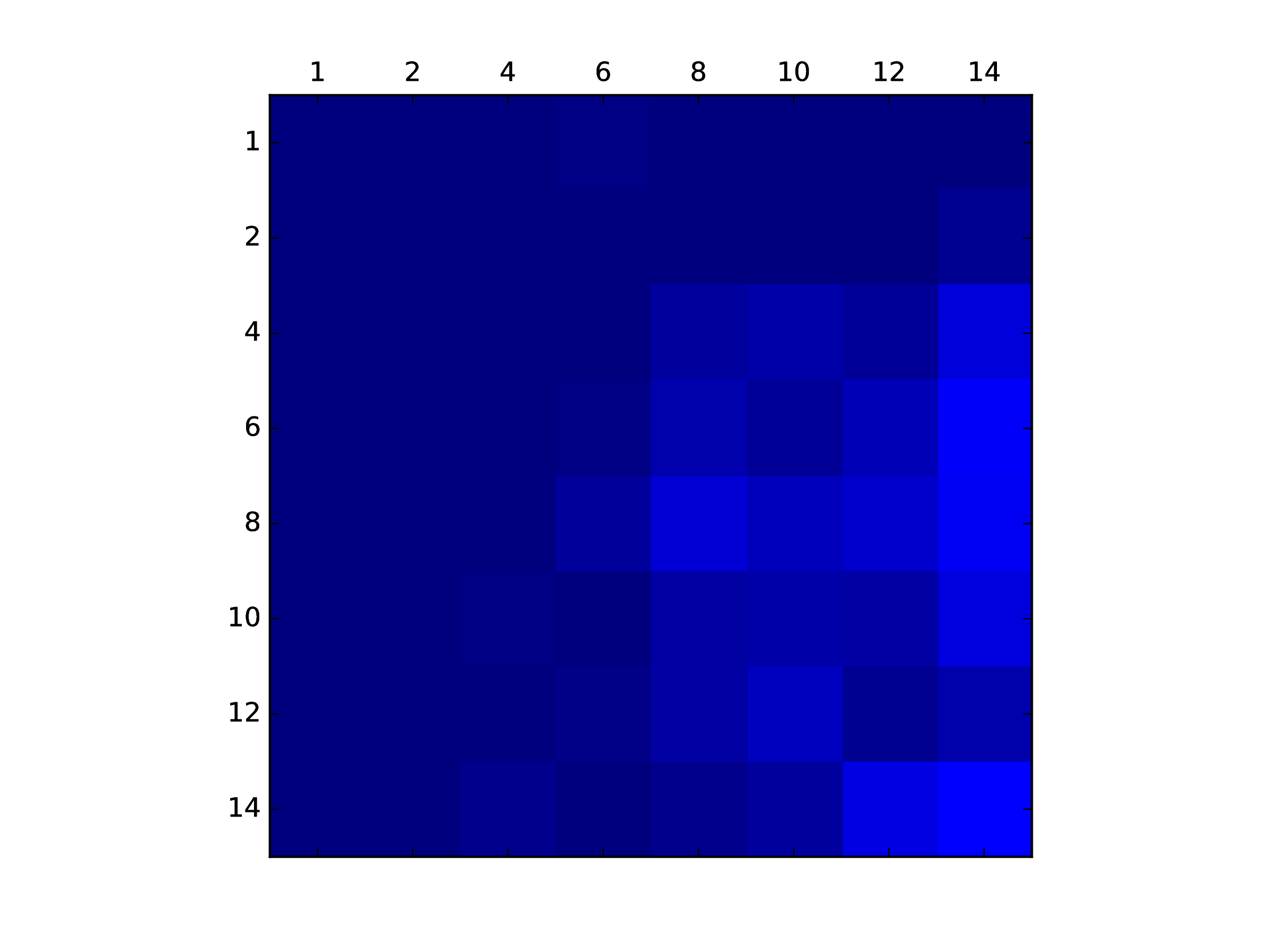}\\
  \end{tabular} % {4.cm 1.2cm 3.7cm 1.cm}
  \caption{Grid simulation over the number of ERPs used in both the average and
    concatenation processes. In the first color map, the first 8 electrodes with
    higher SC were used to perform the simulation. These 8 electrodes, also,
    correspond to the center-frontal area of the scalp. The other 6 color maps show
    the results for the following 6 subsets of 8 electrodes, from higher SC to
    lower. The black square shows the best configuration obtained in the entire simulation.}
  \label{fig:color11100}
\end{figure}

\begin{table*}[]
%  \centering
\hspace{-0.75cm}\begin{tabular}{c|c|c|c|c|c|c|c|}
\cline{2-8}
                                    & \multicolumn{7}{c|}{\textbf{Subsets of Electrodes}}                        \\ \hline
  \multicolumn{1}{|c|}{\textbf{Configuration}} & 0 - 7  & 8 - 15 & 16 - 23 & 24 - 31 & 32 - 39 & 40 - 47 & 48 - 55 \\ \hline
  \multicolumn{1}{|c|}{C1 M1}         & 0.1121 & 0.1044 & 0.0879  & 0.0847  & 0.0972  & 0.0898  & 0.0978  \\ \hline
  \multicolumn{1}{|c|}{C4 M4}         & 0.2371 & 0.1944 & 0.1702  & 0.1355  & 0.1367  & 0.1116  & 0.1108  \\ \hline
  \multicolumn{1}{|c|}{\textbf{C12 M4}}        & \textbf{0.4209} & 0.3093 & \textbf{0.2070}  & 0.1917  & 0.1723  & 0.1053  & 0.1021  \\ \hline
  \multicolumn{1}{|c|}{\textbf{C4 M12}}        & \textbf{0.4818} & 0.4048 & \textbf{0.3813} & 0.3296  & 0.2445  & 0.1626  & 0.1290  \\ \hline
  \multicolumn{1}{|c|}{C12 M12}       & 0.5294 & 0.4564 & 0.4362  & 0.3858  & 0.3616  & 0.1607  & 0.1235  \\ \hline
\end{tabular}
  \caption{Silhouette Coefficients from some configurations of Means (M) and
    Concatenations (C) from figure \ref{fig:color11100}, for different sets of
    electrodes. The columns represent the different subsets of electrodes used
    in each experiment (e.g. 0-7 represents the first 8 electrodes sorted by activity,
    and so on). The rows of the table represent the different object
    configurations used in each experiment. Looking at the configurations C4 M12
  and C12 M4, one can notice difference in the clustering
  quality obtained from each conguration even though the number of ERPs used is
  the same.  Also, the ``degradation'' in the SC caused by taking worse sets of
  electrodes, tends to affect less the configurations with higher Concatenations
  than those with higher Means. This phenomenon is particularly more visible
  between 
  the first and third configurations (highlighted in bold). In the first case
  the SC difference between the two configurations is $\sim 0.006$  while in the
second one is $\sim 0.1743$.}  
  \label{table:mapacolorsincolor}
\end{table*}

As presented in Section \ref{subsec:sys_test}, we have observed that the
clustering quality seems to be correlated 
with the number of segments used to create each object. However, we have explored limited configurations of
the parameters $M$ and $C$. With the aim of analyzing the impact that both parameters have on the clustering
quality, we have performed a more exhaustive grid search.

In this experiment, we have explored configurations from $M = 1$ to $M = 14$ and
from $C = 1$ to $C = 14$. 
Besides, instead of using single electrodes, we have used 8 electrodes in each
experiment. Each 
subset of electrodes has been selected based on the SC obtained in the
experiments presented in Section 
\ref{subsec:source_sim}. Hence, we have sorted the electrodes according
to their median SC, and we 
have created several sets of electrodes (the first set contains the best 8 electrodes, the second set
contains the electrodes in positions 9 to 16, and so on). Figure \ref{fig:color11100} depicts the obtained
results for the first 7 subset of electrodes to show the obtained progression.
Some interesting numerical values of SC that correspond to Figure
\ref{fig:color11100} are presented in Table \ref{table:mapacolorsincolor}, to
show a more detailed version of this experiment.

As we expected, analyzing figure \ref{fig:color11100} and Table
\ref{table:mapacolorsincolor}, it can be noticed that the combination of the
best electrodes provides greater SC than the combination of those electrodes with
lower SC. We have also observed
some differences between the parameters $M$ and $C$, that is, between the average and concatenation
processes. Thus, while the concatenation process always increases the SC, the
average process increases it until $M = 6$. In the table
\ref{table:mapacolorsincolor}, one can see that while the configuration of $M =
4$ and $C = 12$ reports a SC of 0.48, for the best subset of electrodes, the opposite
configuration ($M = 12; C = 4$) reports a SC of 0.42, even though they use the same number of
ERPs per object (48). The fact that the number of ERPs used in the
  concatenation process seems to be more important that the ones used in the
  averaged process can be observed looking at the other configurations of the
table. 

Our hypothesis is that the average reduces the noise 
of the segments but does not provide enough information to generalize 
the method which, in 
this case, is provided by the concatenation. On the other hand, neither the
average nor the concatenation 
processes alone can identify the structure of the objects well enough, probably
due to the combination of noise and differences between P300-ERPs.

\begin{figure}
  \centering
  \includegraphics[width=0.75\textwidth]{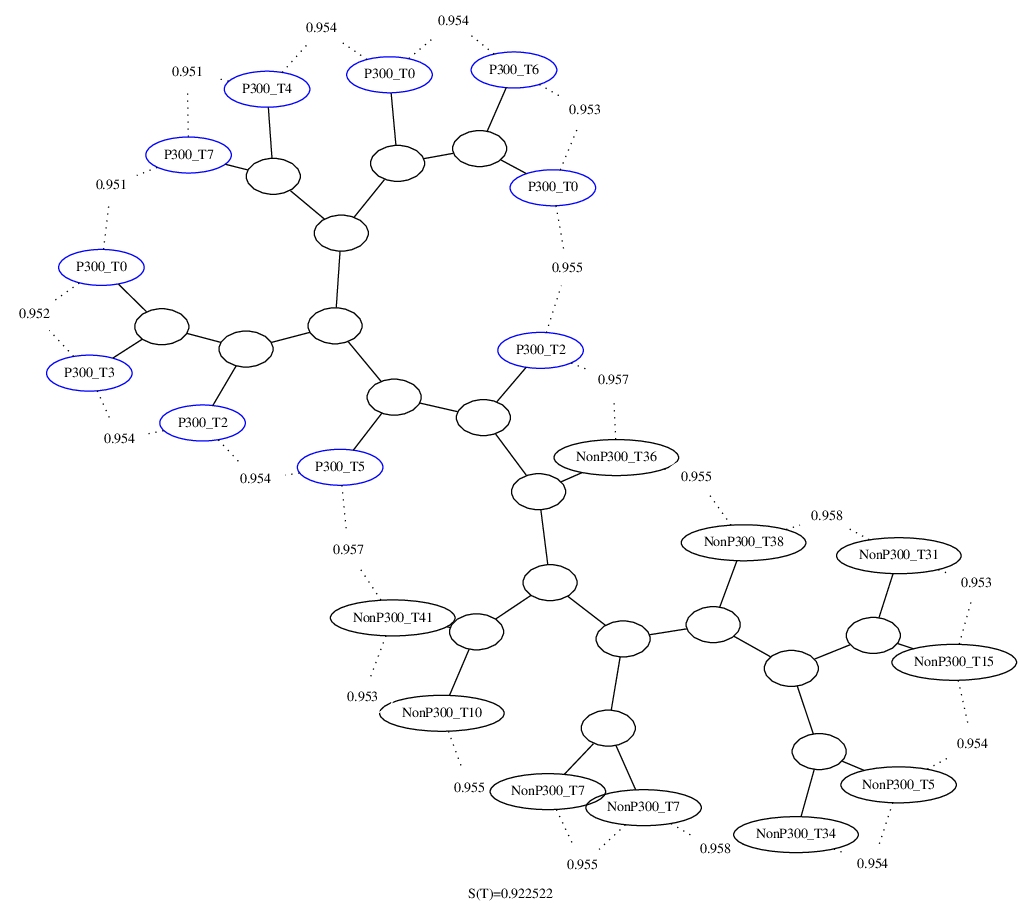}
\caption{Dendrogram (using CompLearn toolkit) obtained from a
  subset of optimized objects. The blue nodes correspond to P300s objects and
  the black ones to non-P300. The Silhouette Coefficient of the dendrogram is
  0.56. Following the scheme of figure \ref{fig:object-definition}, we take a
  configuration for objects creation of $M = 10$ and $C = 12$. This configuration corresponds to the optimal separation of P300 and non-P300 in figure \ref{fig:color11100}. Here, we can see how the separation of P300 and non-P300 from figure \ref{fig:maketree},  where the number of means and concatenations was not optimized, is improved considerably. }
  \label{fig:MQTCbueno}
\end{figure}

\begin{figure}
  \centering
  \includegraphics[width=0.7\textwidth,trim={0cm 1cm 12cm 0.64cm},clip]{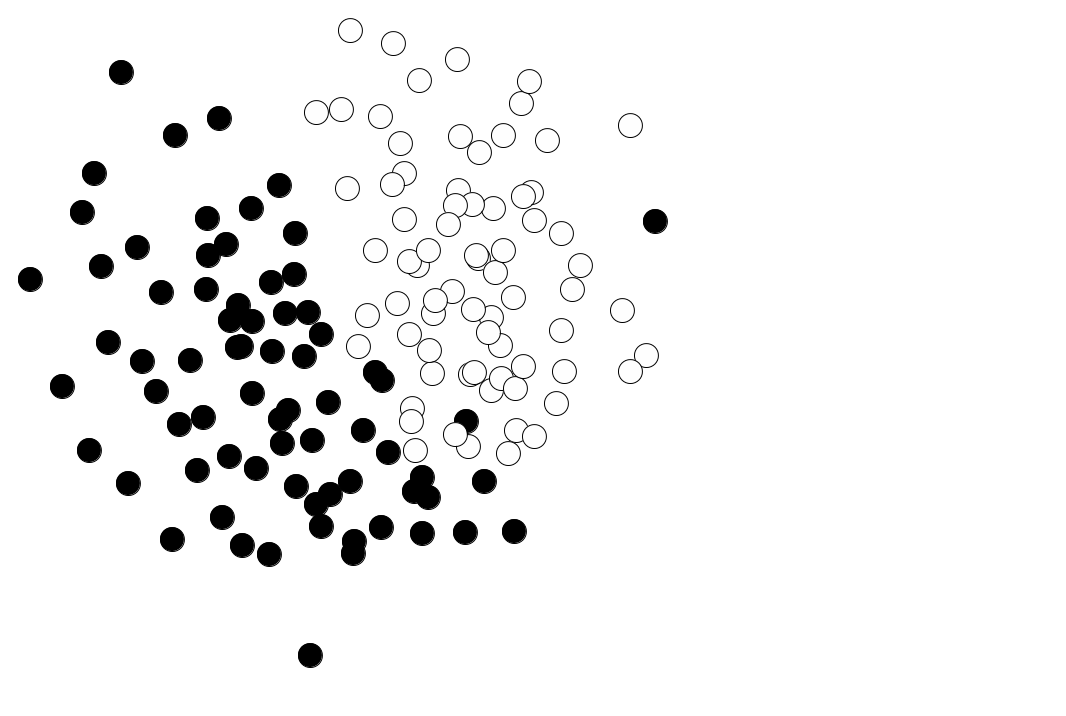}
  \caption{PEx projection from a subset of optimized objects. The white nodes
    correspond to the P300 objects and the black ones to non-P300. The
    Silhouette Coeficient of the projection is 0.36. Following the scheme of
    figure \ref{fig:object-definition}, we take a configuration of $M = 10$ and $C = 12$. This configuration
    corresponds to the optimal separation of P300 and non-P300 in figure
    \ref{fig:color11100}. Here, we
    can see how the separation of P300 and non-P300 from figure \ref{fig:pexsample},  where the
    number of means and concatenations was not optimized, is improved considerably.}
  \label{fig:PExbueno}
\end{figure}

Finally, we have repeated the experiments performed for figures
\ref{fig:maketree} and \ref{fig:pexsample} when the combination of means and
concatenations was not optimized. In these new 
experiments we have taken the best combination of $C$ and $M$, following the
previous experiments of figure \ref{fig:color11100} ($M = 10$ and $C = 12$). As we
show in  
figures \ref{fig:MQTCbueno} and \ref{fig:PExbueno}, the separation between both
clusters is improved compared to the previous experiments. It is
interesting to point out, however, that the SC obtained for both methods, PEx
and CompLearn, differs in magnitude. This is caused by the differences in the measurement of the SC (described in section
\ref{subsec:silcoef}).

\subsection{Validation and generalization of results to extract P300-ERPs}
\label{ssec:validate}

Along this paper, we have performed different experiments to analyze and study
the capabilities of a clustering analysis based on string
compression to identify P300-ERP structure. Now, in this section, we aim
to validate our previous results using a different dataset: BCI Competition
III, dataset II \cite{bci_iii}. Particularly, we have 
performed the same experiments of Sections \ref{subsec:source_sim}, to obtain the best
electrodes of each subject, and \ref{subsec:object_comp}, to achieve a
reasonable P300-ERP clustering, to this dataset. 

\begin{figure}
  \centering
  \begin{tabular}{cc}
  \includegraphics[width=0.51\textwidth]{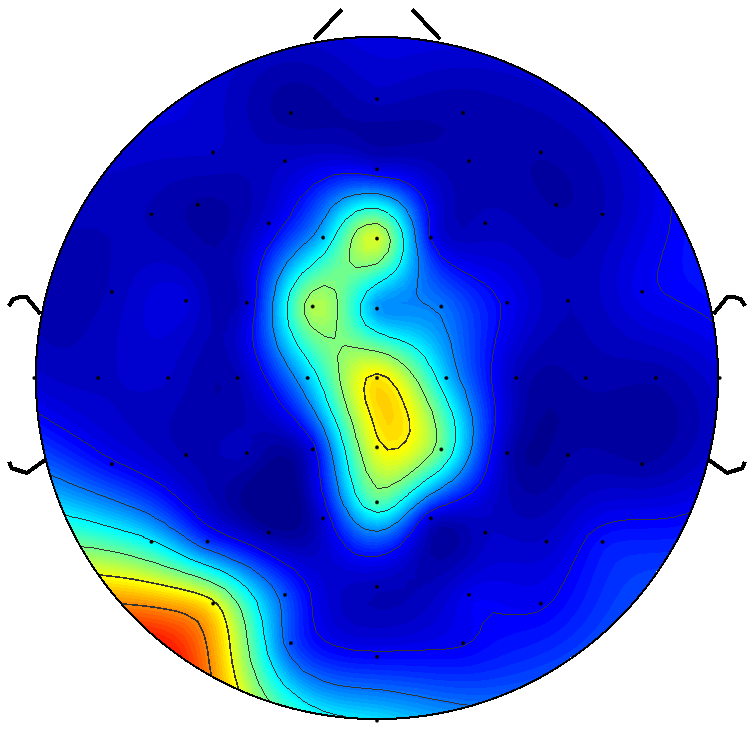} &
  \includegraphics[width=0.54\textwidth]{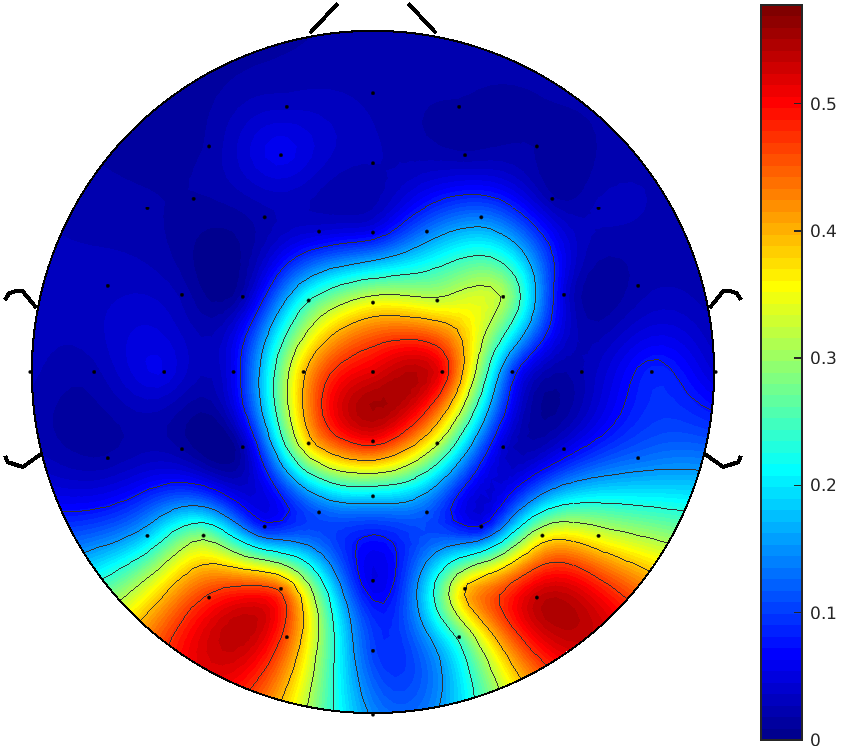} \\
  \end{tabular}
  \caption{Silhouette Coefficient means distribution across the scalp obtained
    for each electrode for the subject A and B, respectively, of the third BCI Competition. Each
    score was calculated from a $C = M = 8$ configuration with CompLearn.}
  \label{fig:Alongcabeza}
\end{figure}

First, following a methodology similar to the one followed in the experiment of figure
\ref{fig:electrodesim}, we have obtained the electrode SC map 
for both subjects A and B (see figure \ref{fig:Alongcabeza}). The configuration
used in this case is the 
same as in the previously mentioned experiment, $C = M = 8$.
In these figures, one can observe that the areas of
activity are more isolated in the third competition than in the experiments of
the second competition. This is given by the signal quality
differences between both competitions. Also, these results (the activity
regions) are consistent with the ones showed by the winners of the competition
in \cite{rakotomamonjy_bci_2008}, which may suggest that our method is capable
of extracting relevant information for a P300-ERP analysis.

\begin{figure}
  \centering
  \begin{tabular}{cc}
    \includegraphics[width=0.42\textwidth]{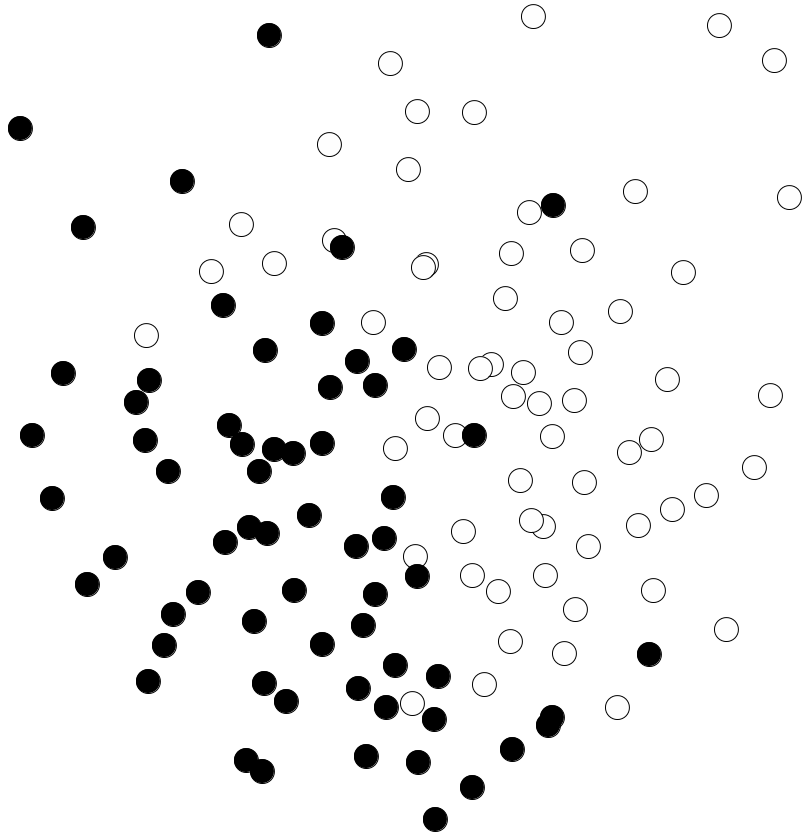} &
    \includegraphics[width=0.44\textwidth]{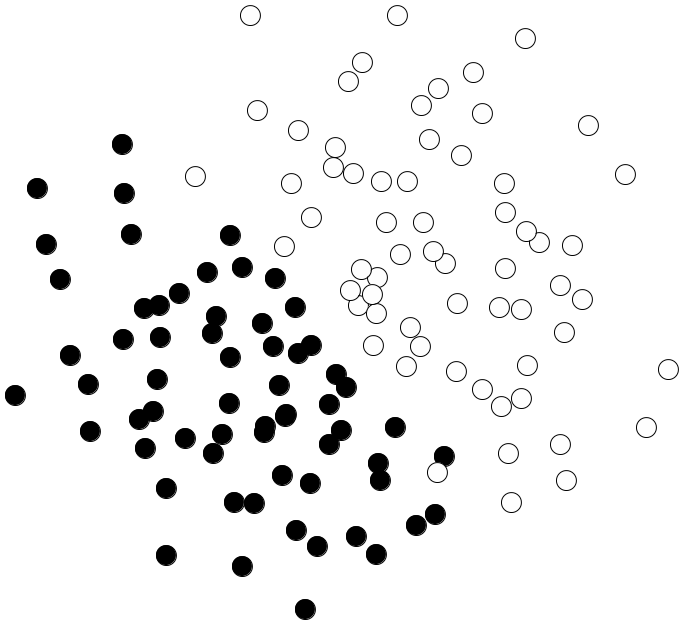} \\
  \end{tabular}
  \caption{PEx projection from Subject A and B, respectively, obtained from the top 3 central-frontal
    subset of optimized objects from the BCI 
    Competition III dataset II. The white nodes
    correspond to the P300 objects and the black ones to non-P300. The
    Silhouette Coeficient of the projections are 0.25 and 0.38, respectively.
    The configuration is $M = 10;  C = 12$.}
    \label{fig:Along}
  \end{figure}

Using the data gathered for the previous experiment (depicted in figure
\ref{fig:Along}), we 
 have combined the best electrodes of the central-frontal zone (as in figure
 \ref{fig:color11100}) into a unique dendrogram, in order to improve the overall
 clustering. For both subjects, we have used the same configuration
used in the experiment of figure \ref{fig:PExbueno}, ($C = 12$
and $M = 10$). In this experiment, we have observed that combining electrodes of
the same zone with a high SC improves the clustering, as we observed using the
same method over the data set of the second BCI Competition. On the other hand,
we have observed that the number of electrodes with high clustering quality is
fairly lower in the BCI Competition III data set, which make sense given the
signal quality differences between both competitions.

%%% Local Variables:
%%% mode: latex
%%% TeX-master: "main"
%%% End:

\section{Discussion}

\label{sec:discussion}

Throughout this work, we have carried out several experiments to test the
capabilities of our NCD-driven clustering methodology. Firstly, we have explored
how varying the parameters of our object-creation 
method affects the clustering performance, preliminary in Section
\ref{subsec:sys_test} and more 
detailed in  Section \ref{subsec:object_comp}.
For our experiments, we have taken two different clustering methods to
double-check our results, and the Silhouette Coefficient as an unbiased quality
measure. The results obtained from these experiments have showed a tendency for
better clustering with higher numbers of P300-ERPs per file.
In figures \ref{fig:pex_comparison} and \ref{fig:dot} this fact is
slightly visible, but it is clear in Figure \ref{fig:color11100}. In the 
experiments of the last figure, we noticed slight differences alternating
between averages and concatenations per file, being the concatenation of
P300-ERPs apparently more important than their average in the long term.
In our opinion, the average of a few P300-ERPs may be sufficient for removing the
majority of noise and variability from an object. On the other hand, the
concatenation of several P300-ERPs in one object, which can be seen as multiple
examples, can 
provide more generalization to the final objects. Besides, the concatenation
process increases the size of objects, which in scenarios where dealing with
small objects, always improves the capabilities of the NCD
\cite{cilibrasi_clustering_2005,granados_improving_2014-1}. 

Finally, the last experiments presented in the paper were targeted to produce a
score for each electrode. These experiments were carried out to compare our
approach with other works in the literature. Congruent results were
obtained first in Section \ref{subsec:source_sim} as a proof of concept, and
second in Section \ref{ssec:validate} using a different dataset to
validate our previous results. In these experiments, we have found remarkable
differences between the activity of the two datasets used in this paper, which
is also congruent with the classification scores published in the website of
each dataset. Thus, while in the first competition five different contestant
achieved 0\% error in the classification, only one contestant achieved an error
below 4\% in the second one (followed by a 9.5 \% error).

\section{Conclusions}
\label{sec:conclusions}
The first objective of this work was to explore whether compression-based
distances could be a useful tool for identifying P300 structure. Derived from
this, the second objective of this work was to develop a method to represent
ERP-based EEG
signals in a manner that compressors are able to identify common structures
between different P300-ERPs. Therefore, we have defined a signal-to-ASCII process to
parse the EEG signal into objects so that they are suitable to be used by a
compression algorithm. The results obtained show that the NCD can be 
successfully applied to the P300 identification using our coding
methodology. We have found consistent results in the P300 activity regions with
other works in the literature. This fact not 
only validates our method of object 
representation, but confirms that compression-based distances are a valid
tool for analyzing P300. 

%%% Local Variables:
%%% mode: latex
%%% TeX-master: "main"
%%% End:

\section*{Acknowledgement}

This work was funded by Spanish project of Ministerio de Econom\'ia y
Competitividad/FEDER TIN2014-54580-R and TIN2017-84452-R
( \url{http://www.mineco.gob.es/} ). The funder had no role in study design, data
collection and analysis, decision to publish, or preparation of the manuscript.
We thank Diana Flores for helping us with the SC measurement of the dendrogram.

\bibliography{mi-thesis,bitacora-ordenada,thesis-aux}

\end{document}